\documentclass[9pt]{article}
\usepackage{extsizes}
\usepackage[super,sort&compress,comma]{natbib} 
\usepackage[version=3]{mhchem}
\usepackage[left=1.5cm, right=1.5cm, top=1.785cm, bottom=2.0cm]{geometry}
\usepackage{balance}
\usepackage{mathptmx}
\usepackage{sectsty}
\usepackage{graphicx} 
\usepackage{lastpage}
\usepackage[format=plain,justification=justified,singlelinecheck=false,font={stretch=1.125,small,sf},labelfont=bf,labelsep=space]{caption}
\usepackage{float}
\usepackage{fancyhdr}
\usepackage{fnpos}
\usepackage[english]{babel}
\addto{\captionsenglish}{%
  
}
\usepackage{array}
\usepackage{droidsans}
\usepackage{charter}
\usepackage[T1]{fontenc}
\usepackage[usenames,dvipsnames]{xcolor}
\usepackage{setspace}
\usepackage[compact]{titlesec}
\usepackage{hyperref}
\usepackage{appendix}

\usepackage{epstopdf}

\definecolor{cream}{RGB}{222,217,201}
\usepackage{stfloats}
\usepackage{subcaption}
\usepackage{amssymb}
\usepackage{booktabs}
\usepackage[normalem]{ulem}
\useunder{\uline}{\ul}{}

\normalfont\upshape
\rmfamily
\sectionfont{\sffamily\Large}
\subsectionfont{\normalsize}
\subsubsectionfont{\bf}

\addto{\captionsenglish}{%
  
}

\title{A Cartesian Encoding Graph Neural Network for Crystal Structures Property Prediction: Application to Thermal Ellipsoid Estimation}

\date{}

\begin{document}

\maketitle
\vspace{-1.5cm}
\begin{center}

\begin{minipage}{.8\textwidth}
    \begin{center}

    \noindent\large{Àlex Solé\textit{$^{a,b}$}, Albert Mosella-Montoro\textit{$^{a}$}, Joan Cardona\textit{$^{b}$}, Silvia Gómez-Coca$^{\ast}$\textit{$^{b}$}, Daniel Aravena$^{\ast}$\textit{$^{c}$}, Eliseo Ruiz$^{\ast}$\textit{$^b$} and Javier Ruiz-Hidalgo$^{\ast}$\textit{$^{a}$}}
    \end{center}
\end{minipage}

\begin{minipage}{0.5\linewidth}
    \vspace{0.5cm}
\begin{center}

\small{
\textit{$^{a}$~Image Processing Group - Signal Theory and Communications Department, Universitat Politècnica de Catalunya, Barcelona, Spain; E-mail: j.ruiz@upc.edu}

\textit{$^{b}$~Inorganic and Organic Chemistry Department and Institute of Theoretical and Computational Chemistry, Universitat de Barcelona, Barcelona, Spain; E-mail: silvia.gomez@qi.ub.es, eliseo.ruiz@qi.ub.edu}

\textit{$^{c}$~Materials Chemistry Department, Faculty of Chemistry and Biology, Universidad de Santiago de Chile, Santiago, Chile. E-mail: daniel.aravena.p@usach.cl}

}
\end{center}
\end{minipage}

\end{center}


\begin{abstract}
In the diffraction resolution of crystal structures, the thermal ellipsoids are a critical parameter that is usually more difficult to determine than atomic positions. 
These ellipsoids are quantified through the Anisotropic Displacement Parameters (ADPs), which provide critical insights into atomic vibrations within crystalline structures. 
ADPs reflect the thermal behaviour and structural properties of crystal structures. 
However, traditional methods to compute ADPs are computationally intensive. This paper presents CartNet, a novel graph neural network (GNN) architecture designed to predict properties of crystal structures efficiently by encoding the atomic structural geometry to the cartesian axes and the temperature of the crystal structure.
Additionally, CartNet employs a neighbour equalization technique for message passing to help emphasise the covalent and contact interactions, and a novel Cholesky-based head to ensure valid ADP predictions.
Furthermore, a rotational SO(3) data augmentation technique has been proposed during the training phase to generalize unseen rotations.
To corroborate such procedure, an ADP dataset with over 200,000 experimental crystal structures from the Cambridge Structural Database (CSD) has been curated. 
The model significantly reduces computational costs and outperforms existing previously resported methods in ADP prediction by 10.87\%, while demonstrating a 34.77\% improvement over the tested theoretical computation methods.
Moreover, we have employed CartNet for other already known datasets that included different material properties, such as formation energy, band gap, total energy, energy above the convex hull, bulk moduli, and shear moduli. 
The proposed architecture outperformed previously reported methods by 7.71\% in the Jarvis Dataset and 13.16\% in the Materials Project Dataset, proving CarNet's capability to achieve state-of-the-art results in several tasks.
Project website with online demo available at: \href{https://www.ee.ub.edu/cartnet}{https://www.ee.ub.edu/cartnet}
\end{abstract}



\section{Introduction}

The Anisotropic Displacement Parameters (ADP)~\cite{Cruickshank:adp} represent a three-dimensional ellipsoid of atomic displacements within a crystal lattice due to thermal vibrations. 
These ellipsoids are fundamental for understanding the anisotropic nature of atomic movements and the dynamic behaviour of materials at the atomic scale. 
They provide critical insights into the structural and thermal properties of materials, influencing the interpretation of experimental data from techniques such as single crystal X-ray diffraction and neutron scattering. 
Accurate representation of ADPs is essential for constructing precise structural models and understanding the physical behaviour of materials under varying thermal conditions.

The ADPs are particularly valuable in crystallography, as they assist in identifying determination issues such as disorder or twinning. Several studies have shown that ADPs can also be utilized to predict thermal motion, as well as translational and vibrational frequencies~\cite{Cruickshank:adp, Cruickshank:a01881}. 
Furthermore, as demonstrated in previous research, thermal properties like heat capacity ($C_v$)~\cite{doi:10.1021/jp062953a, doi.org/10.1002/anie.200906780} and vibrational entropy~\cite{e47cd8a436ae41018cd8a02c1584f934,Cruickshank:a01882} are directly linked to ADPs. 
Additionally, ADPs have been related to the thermal expansion of crystal structures~\cite{doi:10.1126/science.1151442,PhysRevB.48.3156}, making them especially useful for identifying materials with negative thermal expansion.
From an experimental point of view, inconsistencies in atomic positions are often easily spotted using chemical intuition when solving a crystal structure, allowing for straightforward visual identification of errors. 
However, such intuitive assessments are not as straightforward  when it comes to thermal ellipsoids, making visual detection of discrepancies more challenging. 
As a result, some database structures exhibit ellipsoids with seemingly anomalous sizes and orientations. 
Developing theoretical methods that enable quick evaluation of these ellipsoids can thus serve as valuable tools for facilitating accurate structural determination from diffraction data.

From a theoretical perspective, ADPs can be calculated using periodic electronic structure calculations to obtain the vibrational frequencies based on the harmonic approximation~\cite{doi:10.1021/acs.accounts.7b00067, 10.1039/C5CE01219H}. 
This method requires the numerical calculation of forces at displaced geometries for all atomic positions in solid-state systems. 
Thus, these calculations are computationally expensive and time-consuming, often creating a bottleneck in the process. 
Our contribution is particularly significant in this context, as deep learning methods based on graph neural networks can dramatically reduce the computation time required, providing a more efficient alternative to traditional approaches.

\begin{figure}[htb]
    \centering
    \includegraphics[width=\textwidth]{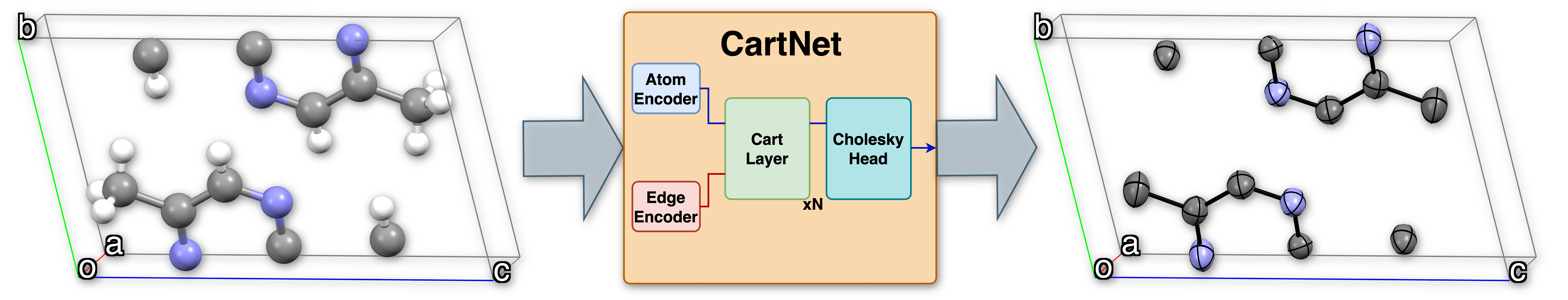}
    \caption{Schematic of the CartNet graph neural network architecture for a \textit{5,5'-dimethyl-2,2'-bipyrazine} crystal structure (CSD Refcode: ETIDEQ). The model predicts the ADPs for all non-hydrogen atoms based on the positions of atoms within the unit cell. The architecture separately encodes atomic and edge information using dedicated encoders. This information is then aggregated through N iterations of message-passing via the CartLayer. Finally, the Cholesky Head ensures the output matrix is symmetric and positive-definite, generating a valid ADP matrix. White, light purple, and grey colours represent hydrogen, nitrogen, and carbon atoms, respectively. The parallelepiped represents the unit cell, and the red, green, and blue lines correspond to the a, b, and c unit cell axes. }

    \label{fig:architecture}
\end{figure}

Figure \ref{fig:architecture} shows the full pipeline proposed for this work, named CartNet.
Our presented network uses a graph representation of the crystal structure and, through a set of learnable encodings and geometrical operations, predicts atomic or material properties

Moreover, predicting ADPs presents a unique challenge and opportunity in the field of machine learning for crystal structure property prediction. 
Unlike most tasks typically explored in the literature, such as formation energy, band gap, or total energy, which are unaffected by rotation, predicting ADPs requires models to be sensitive to rotational orientation. 

In this paper, we make the following contributions:

\begin{enumerate}
    \item \textbf{ADP Dataset:} We present a meticulously curated dataset of ADPs with over 200,000 experimental crystal structures from the Cambridge Structural Database (CSD)~\cite{csd_webpage}.
    The ADPs are both temperature-dependent and rotation-equivariant, offering a robust foundation for exploring the necessity and impact of rotation-equivariant architectures in crystal modelling.
    This necessitates the use of architectures that can handle properties dependent on direction, opening the door to designing networks capable of processing rotationally dependent features. 
    See Section \ref{sec:dataset} for a detailed description of the dataset.

    \item \textbf{CartNet Architecture:} We introduce an architecture, CartNet, capable of accurately predicting material properties based on the geometry of a crystal structure.
    The key contributions of this model are as follows:

    \begin{enumerate}
        \item \textbf{Geometry and Temperature Encoding:} We propose a feature descriptor that efficiently encodes the complete 3D geometry referenced to the Cartesian axis and adaptively fuse other input attributes such as temperature.
        Since the geometry is ancored to the Cartesian reference axes, there is no need to encode the unit cell.
        %
        Cell-less encoding enables the accurate prediction of the crystal's ADP orientation regardless of cell size.

         \item \textbf{Neighbour Equalization:} We present a neighbour equalization technique designed to address the exponential increase in neighbours over distance, thereby enhancing the model's ability to detect various types of bonds and force interactions between atoms.
        This technique improves the model's sensitivity to different interaction ranges, ensuring a more precise representation of atomic environments.

        \item \textbf{Cholesky Head:} We introduce a output layer based on Cholesky decomposition, which guarantees that the model produces positive definite matrices, an essential mathematical property requirement for valid ADP predictions.
    
    \end{enumerate}

    \item \textbf{Rotation SO(3) Augmentation:} We propose a data augmentation technique for both input features and output ADPs to facilitate the creation of SO(3) rotation-equivariant representations.
    This technique enables the model to learn rotational equivariance without requiring specific layers to enforce it, thereby simplifying the overall architecture and reducing the number of trainable parameters.

\end{enumerate}

These contributions allowed our model to outperform previously reported methods by 8.85\% in Jarvis Dataset~\cite{choudhary2020joint} and 15.5\%  in the Materials Project Dataset~\cite{materials_project}, vide infra.

Furthermore, in the ADP dataset examined, CartNet demonstrated a 10.87\% improvement over other previously reported methods and a 34.77\% improvement over the tested low-level Generalized Gradient Approximation (GGA) Density Functional Theory (DFT) calculations with simple dispersion corrections.

\section{Background and Related Work}

\subsection{Thermal Ellipsoids}

The Anisotropic Displacement Parameters (ADPs) represent the magnitudes and directions of atomic thermal vibrations within crystal structures.
ADPs encapsulate the statistical probability distribution of the position of the atoms resulting from thermal vibrations.
Typically, ADPs are graphically represented using ellipsoids within the ORTEP visualization~\cite{Ortep} that depicts a 50\% probability contour indicating the likelihood of finding the atom within the ellipsoid's bounds.

Mathematically, ADPs are represented as a three-dimensional tensor ($3 \times 3$ matrix), which functions as a covariance matrix for a three-dimensional Gaussian distribution.
The covariance matrix comprises variance elements along its diagonal for each $X$, $Y$ and $Z$ axis and covariance elements off the diagonal, illustrating the relationships between the axes. 
The covariance matrix $\mathbf{U}$ for a three-dimensional Gaussian distribution can be expressed by the Equation (\ref{eq:cov_matrix}).

\begin{equation}
\centering
\mathbf{U} = 
\begin{bmatrix}
    \text{Var}(X) & \text{Cov}(X, Y) & \text{Cov}(X, Z) \\
    \text{Cov}(Y, X) & \text{Var}(Y) & \text{Cov}(Y, Z) \\
    \text{Cov}(Z, X) & \text{Cov}(Z, Y) & \text{Var}(Z)
\end{bmatrix}
\label{eq:cov_matrix}
\end{equation}

Covariance matrices exhibit distinctive mathematical properties: they are symmetric, $\mathbf{U} = \mathbf{U}^T$, and positive semidefinite, which entails that all matrix eigenvalues are non-negative.

\begin{figure}[htbp]
    \centering
    \includegraphics[width=0.3\linewidth]{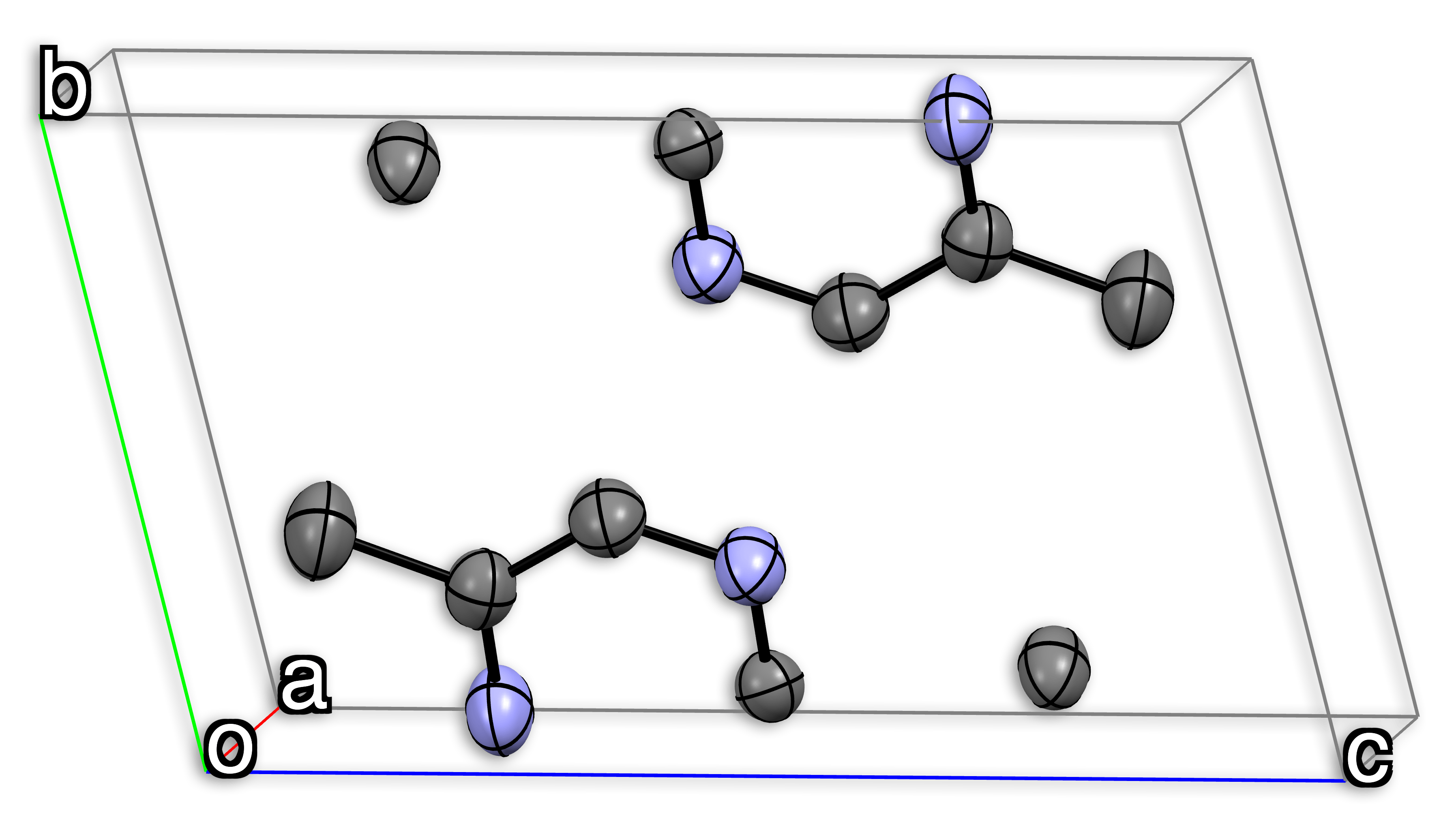}
    \caption{Thermal ellipsoids ORTEP representations from experimental ADPs of an \textit{5,5'-dimethyl-2,2'-bipyrazine} crystal structure (CSD Refcode: ETIDEQ). Light purple, and grey colours represent nitrogen, and carbon atoms, respectively. Hydrogen atoms have been omitted. The parallelepiped represents the unit cell, and the red, green, and blue lines correspond to the a, b, and c unit cell axis.}
    \label{fig:aspirin}
\end{figure}

Figure \ref{fig:aspirin} shows an example with the graphical representation of the ADPs of an \textit{5,5'-dimethyl-2,2'-bipyrazine} unit cell. 
In the Figure, each atom is depicted using the thermal ellipsoid representations of the ADPs obtained experimentally by X-ray difraction.

ADPs can be calculated theoretically by computing the dynamical matrix of the crystal structure~\cite{phonon_equation,phonopy_adp}.
The dynamical matrix is a fundamental concept in solid-state physics used to describe the vibrations of atoms in a crystal lattice, known as phonons. 
It is a mathematical construct that captures how atoms interact with each other when they are slightly displaced from their equilibrium positions.
The dynamical matrix $\mathbf{D}(\mathbf{q})$ is computed at each point $\mathbf{q}$ in the Brillouin zone.
The Brillouin zone represents the fundamental region in reciprocal space that contains all the unique wave vectors necessary to describe a crystal structure's physical properties.
The eigenvalues of the dynamical matrix represent the phonon's frequencies $\omega_\nu(\mathbf{q})$, also known as phonon modes.

ADPs can be calculated by integrating all the phonon modes for all the $\mathbf{q}$ points using the Equation (\ref{eq:u-phono}).

\begin{equation}
    \mathbf{U}(j, T)=\frac{\hslash}{2 N m_j} \sum_{\mathbf{q}, \nu} \frac{\left(1+2 n_\nu(\mathbf{q}, T)\right)}{\omega_\nu(\mathbf{q})} \mathbf{e}_\nu(j, \mathbf{q}) \otimes \mathbf{e}_\nu^*(j, \mathbf{q})
\label{eq:u-phono}
\end{equation}

Where $j$ is the atom, $T$ is the temperature, $\hslash$ is the reduced Planck constant, $N$ are the number of unit cells, $m_j$ is the atomic mass, $\mathbf{e_\nu}$ are the eigenvectors of the dynamical matrix, $\otimes$ is the outer product, and $n_\nu$ is the phonon population.
The phonon population can be described by Equation (\ref{eq:phono-population}).

\begin{equation}
n_\nu(\mathbf{q}, T)=\frac{1}{\exp \left(\hslash \omega_\nu(\mathbf{q}) / \mathrm{k}_{\mathrm{B}} T\right)-1}
\label{eq:phono-population}
\end{equation}

Where $k_B$ is the Boltzmann constant.

This method has the limitation that it is extremely time-consuming since it needs a different Density Functional Theory (DFT) calculation for each of the $\mathbf{q}$ points.
High-performance computing (HPC) clusters are often needed since DFT calculations can be computationally demanding, even for a single crystal structure.

\subsection{Crystal Structure Property Prediction with Graph Neural Networks}

GNNs~\cite{4700287} have been the most widely adopted neural network architecture for modelling data with complex relational structures like molecules~\cite{gilmer2017neural,ramakrishnan2014quantum,zhou2023unimol} or crystal structures~\cite{matformer, potnet, conformer}.
Unlike traditional neural networks, which typically operate on fixed-size grid-like structures, GNNs are designed to work on graph-structured data, where entities are represented as nodes and relationships between them as edges. 

In the case of molecules, this ability to use graphs to model the system is especially useful, since a molecule can be naturally modelled as a graph of atoms.
However, how atoms are interconnected by edges in the graph, denoted as neighbourhood, is not trivial and becomes a fundamental step to achieve good representations using GNNs.

While GNN and message-passing mechanisms had been introduced previously, Gilmer et al.~\cite{gilmer2017neural} popularized a unified message-passing framework specifically for quantum chemistry predictions, leveraging iterative information exchange between nodes (atoms) and edges (bonds) to capture local molecular interactions. 
This approach laid the groundwork for subsequent models that further enhance predictive accuracy for molecular properties.
DimeNet~\cite{gasteiger_dimenet_2020} and its successor, DimeNet++~\cite{gasteiger_dimenetpp_2020}, build a GNN upon this concept by incorporating directional message passing between pairs of interactions, allowing these models to capture angular dependencies and long-range interactions within molecules more effectively than traditional message-passing methods using chemical bonds.
This advancement significantly improves the accuracy of predicting molecular properties, such as quantum mechanical characteristics. 
Notably, both DimeNet and DimeNet++ are rotation-invariant, ensuring their predictions remain consistent regardless of the molecule's orientation, which is crucial for accurate modelling in diverse molecular environments.
MACE~\cite{batatia2022mace} proposed an equivariant message-passing approach that incorporates advanced three-dimensional geometric information, ensuring rotational and translational symmetries for robust generalization across diverse molecular configurations. 
By enforcing rotational and translational symmetries, MACE provides robust generalization across diverse molecular configurations. 
Similarly, TensorNet~\cite{simeon2024tensornet} utilizes tensor-factorization techniques to incorporate higher-order interactions efficiently, capturing subtle quantum effects and long-range correlations. 
These approaches represent a significant evolution in molecular modeling, surpassing earlier methods in both accuracy and computational efficiency.

On the other hand, crystal structures consist of an assembly of atoms, ions or molecules that are ordered in a symmetric way. 
The formed symmetric pattern is repeated in the three spatial dimensions.  
We represent these structures using their smallest repeating unit, the unit cell, defined by a $3 \times 3$ lattice matrix encoding the three vectors that describe its geometry.
Similar GNN models have been extensively applied to crystal property prediction, showcasing their adaptability to the unique challenges of materials science.
One critical adaptation in this domain involves modifying the neighbourhood definition to account for periodic boundary conditions (PBCs).
PBCs are essential for modelling the infinite nature of crystalline materials. 
They treat the border of the simulation cell as if they were connected to the opposite border, thus creating a continuous, repeating structure.
This modification ensures that the model accurately captures interactions across the boundaries of the material.

Matformer~\cite{matformer} introduces a transformer-like GNN architecture designed explicitly for material property prediction. 
It proposes adding self-loops for each atom to encode the cell dimensions and the PBC radius neighbourhood, thereby enhancing the model's ability to represent the material's structure accurately.
PotNet~\cite{potnet} advances this approach by presenting a GNN with dual-neighbourhood strategy.
In addition to the PBC radius neighbourhood, PotNet approximates the infinite summation of interactions for every pair of atoms in the crystal, providing a more comprehensive representation of the material's properties.
Yang et al.~\cite{conformer} present two GNN architectures tailored for this task, the iConformer and eConformer.
The iConformer refines the cell to create a unique representation and uses the angle between the cell's axis and the Cartesian vector between atom pairs in the PBC radius neighbourhood to produce an invariant representation.
In contrast, the eConformer introduces a rotation-equivariant model using Tensor Products~\cite{geiger2022e3nn}, relying solely on Cartesian information to maintain consistency across different orientations.

Regarding the specific problem of ADPs, current state-of-the-art methods exhibit several limitations. 
One of them is that most approaches rely solely on the distance between atoms to create a rotationally invariant representation. 
Since ADPs are expressed relative to the system's Cartesian axes ($XYZ$), using only distance does not provide the necessary references. 
Models that depend exclusively on distance are unable to differentiate between the variances along the axes ($\text{Var}(X)$, $\text{Var}(Y)$, or $\text{Var}(Z)$), often resulting in spherical ellipsoids that fail to capture the true anisotropy of the ADPs.

Another limitation of current state-of-the-art methods arises from the requirement for the lattice matrix.
This is particularly problematic because multiple lattice matrices can represent the same crystal structure, as discussed in iComformer~\cite{conformer}. 
Although iComformer proposes a solution by creating a unique representation for each cell, several challenges remain when using this approach.
The unique representation proposed by iComformer faces a border case when all three axes of the lattice matrix have the same length ($a = b = c$). 
Since iComformer's approach is based on the assumption that $a < b < c$, when the lattice matrix is in this degenerate case, the model is unable to differentiate between $\text{Var}(X)$, $\text{Var}(Y)$, or $\text{Var}(Z)$, once again resulting in spherical ellipsoids.

Our proposed solution is a cell-less architecture that directly encodes the complete 3D geometry referenced to the Cartesian axes instead of only Euclidean distance-based approaches.
This approach avoids the issues of border case scenarios and cell-based overfitting, as it relies entirely on the Cartesian space.
Furthermore, it allows the inclusion of PBCs, providing a more robust and accurate representation of the anisotropy in ADPs without the need for a lattice matrix.

\section{Dataset}
\label{sec:dataset}

\begin{figure}[htb]
\centering
\begin{minipage}{.45\textwidth}
  \centering
  \includegraphics[width=0.8\linewidth]{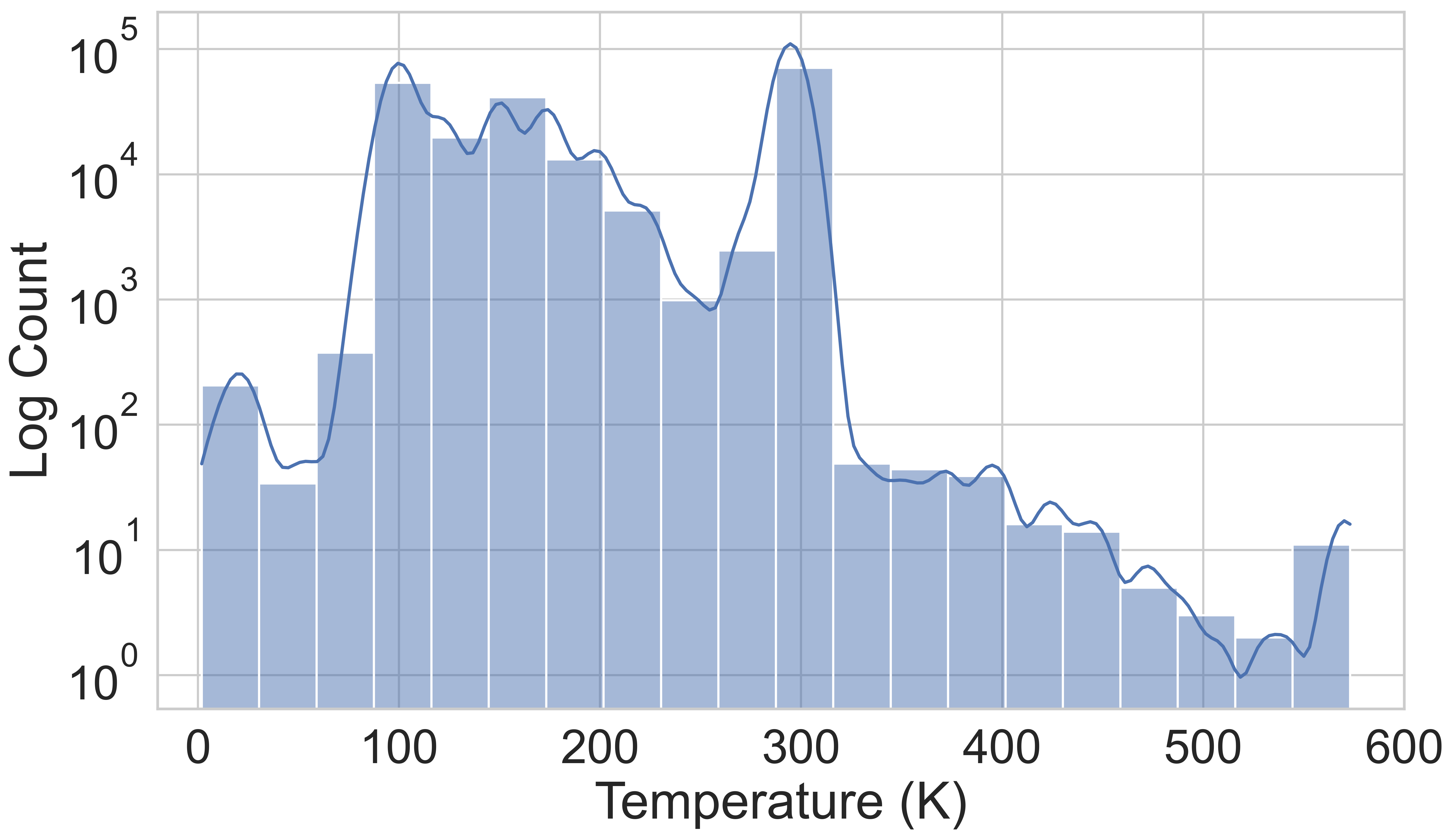}
  \captionof{figure}{Histogram showing the number of crystal structures within each temperature range in the ADP dataset, displayed on a logarithmic scale on the y axis.}
  \label{fig:hist_temp}
\end{minipage}%
\hfill
\begin{minipage}{0.45\textwidth}
  \centering
  \includegraphics[width=0.8\linewidth]{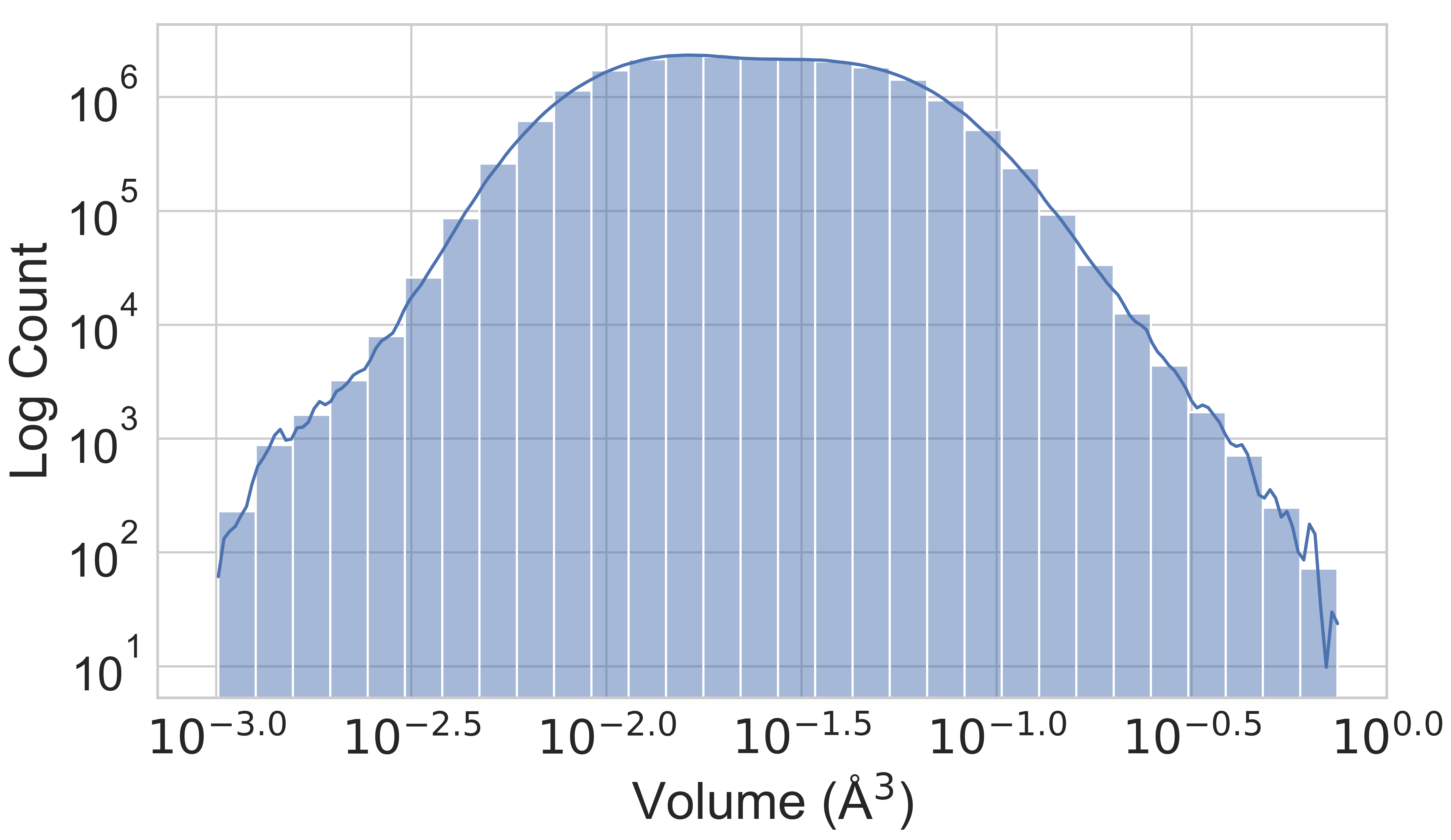}
  \captionof{figure}{Histogram illustrating the number of atoms within each ADP volume range in the ADP dataset, presented on a logarithmic scale on both axes.}
  \label{fig:uequi}
\end{minipage}
    \centering
    \includegraphics[width=0.6\linewidth]{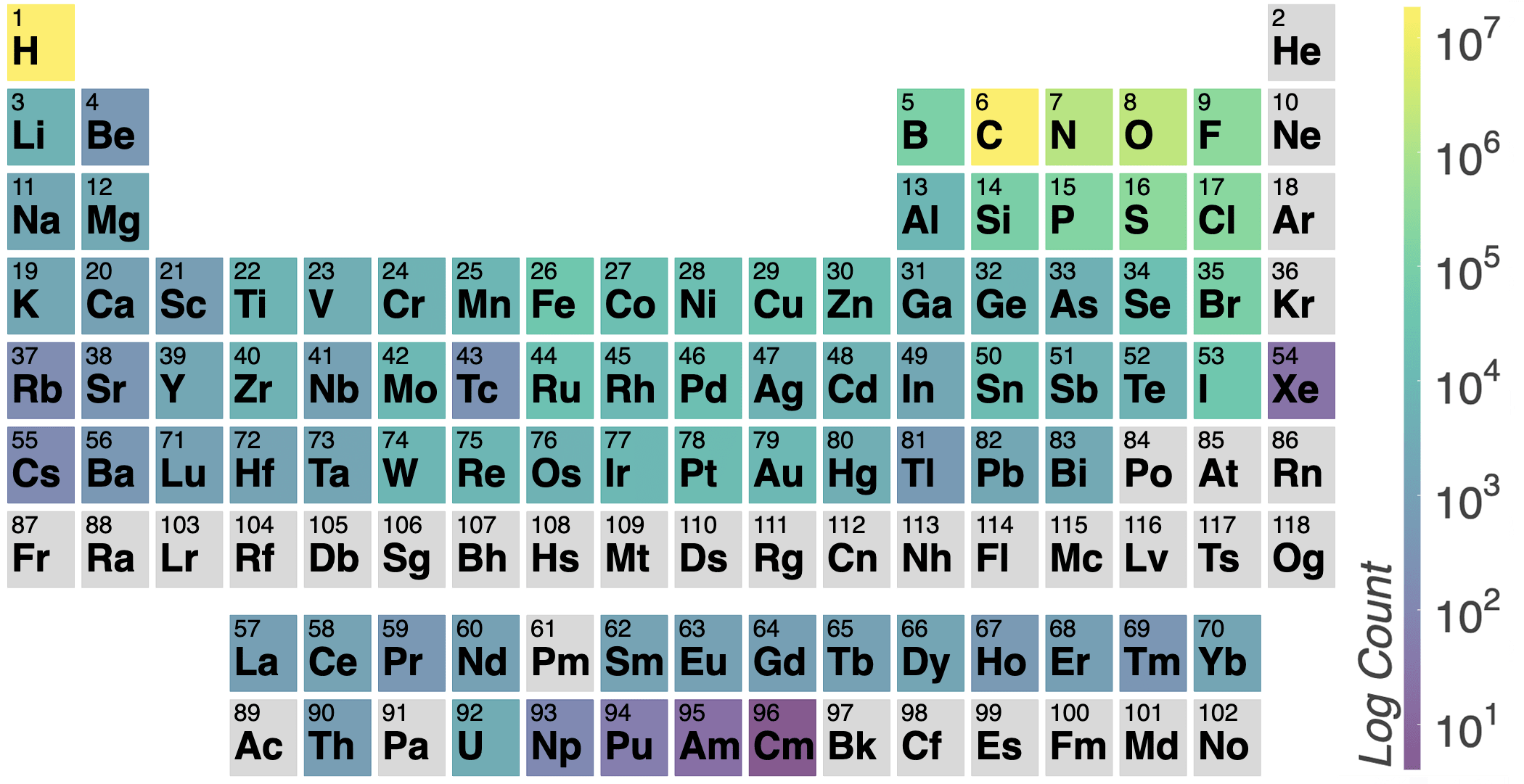}
    \caption{Heatmap illustrating the number of atoms per element in the ADP dataset. Lighter colours represent higher counts, while darker colours indicate lower counts. The colour scale is logarithmic. }
    \label{fig:hist_z}
\end{figure}

The ADP Dataset of crystal structures has been created to facilitate the study of atomic thermal vibrations. 
This dataset is derived from the Cambridge Structural Database (CSD)~\cite{csd_webpage}.

The ADP Dataset was meticulously curated from the CSD's built-in ADP subset, applying several filtering criteria to ensure its quality and reliability.
Firstly, only structures possessing 3D coordinates for all atoms and anisotropic thermal displacements for all non-hydrogen atoms were selected.
Only non-polymeric crystal structures with only one type of molecule in the unit cell were considered, to avoid dispersion and errors due to solvent molecules and counterions.
Structures with an $R$-factor less than $5\%$ ($R < 5\%$), free from errors and disorders, and site occupancy of 1 for all atoms were included to maintain the dataset's integrity.
Structures reported at non-standard pressures were also discarded.

Additionally, temperature/pressure data in the CSD can sometimes be incomplete or missing.
For pressure, it was observed that in many cases the pressure was recorded only in the remarks field rather than in the dedicated pressure field, so structures containing any kind of remarks were discarded to avoid extensive text parsing.
For temperature, if the CSD Python API did not provide a value, the corresponding CIF files and experimental notes were cross-checked to verify consistency.
Structures lacking reliable temperature information were excluded from the dataset.

Several additional filtering criteria were applied to ensure the quality and reliability of the ADPs.
While the 3D positions of the hydrogen are saved, their thermal ellipsoids are not included, as these are often isotropic or undefined in the CSD.
Structures containing any non-hydrogen atom with negative or zero eigenvalues were excluded.
Structures with any ellipsoids with eigenvalue ($\lambda$) ratios $\lambda_{\text{max}} / \lambda_{\text{min}} < 8$ were excluded to avoid poorly defined or flat ellipsoids.
The ratio between the ellipsoid volume and the volume of a sphere with the covalent radius~\cite{cov_rad}, $\text{Vol} / \text{Vol}_{\text{cov}}$, was also used as a filtering criterion.
Structures were discarded if any ellipsoid had a volume greater than $1.25\,\text{\AA}^3$ or if $\text{Vol} / \text{Vol}_{\text{cov}} > 0.35$.
Furthermore, structures with any ellipsoid exhibiting a volume ratio $\text{Vol} / \text{Vol}_{\text{cov}} < 10^{-4}$ at temperatures above $150\,\text{K}$ were also excluded to remove ellipsoids that were either too small or insufficiently defined.

At the end, $208{,}042$ crystal structures from the CSD met the criteria outlined, resulting in an average number of $194.2$ atoms and $105.95$ ADPs per crystal structure.
Since the ADPs of hydrogen atoms are not considered, the dataset has more atoms than ADPs per crystal.

Figure \ref{fig:hist_temp} illustrates the temperature distribution across the curated dataset. 
Even though the dataset spans a wide range of temperatures, from $2\,\text{K}$ to $573\,\text{K}$, most structures rely on the range from $100\,\text{K}$ to $300\,\text{K}$ since this range is the most commonly studied by diffractometry.
Figure \ref{fig:uequi} displays the distribution of the ADP's volumes used for this dataset.
The same behaviour applies to the volumes, since the charts' extremes are under-represented in our data. 
Section~\ref{sec:res_adp} discusses the impact of this imbalance of the data on the prediction performance of the proposed model.
Figure \ref{fig:hist_z} presents a heatmap of the atomic numbers included in this dataset.
The dataset encompasses a wide range of atomic numbers, reflecting a diverse set of elements. 
It excludes some noble gases and most of the radioactive elements, except for some radioactive actinoids.
This diversity is crucial for ensuring that the dataset can support the development of generalised models capable of predicting ADPs across a broad spectrum of chemical compositions.

In this study, the dataset was randomly split into $162{,}270$ training, $22{,}219$ validation, and $23{,}553$ test crystal structures.
To ensure the integrity of these splits, we verified that all atom types, temperature ranges, and volume ranges present in the validation and test sets are also represented in the training set. 
This precaution was taken to prevent the model from encountering unseen data during validation and testing, which could otherwise lead to an inaccurate assessment of its performance.

Additionally, we ensured that repeated crystal structures with different temperatures or distinct CSD entries were kept together within the same split. 
This restriction was made to avoid any situation in which the model might be exposed to test or validation samples during the training phase, which could compromise the evaluation by introducing data leakage. 
By maintaining these strict controls on the dataset splits, we ensured that the training, validation, and test sets were independent.
The splits used for this work are also publicly available to facilitate reproducibility.

\section{Methodology}

\subsection{Model architecture}

Our proposed architecture, CartNet, efficiently encodes the geometry and any other relevant information of the crystal structures. 
The geometrical structure is encoded in the edges via cartesian unit directions, and the input information is encoded along the nodes. 
In the case of ADPs, the input information is the temperature and the atomic number of each atom in the structure. 
The atom and edge information are then iteratively aggregated through the CartLayers, a process that is repeated four times to create a final vector representation for each atom. This final output is then processed through a specific head to predict the final property. 
In the case of ADPs, the Cholesky Head is used to produce mathematically valid ADPs.
The complete architecture of CartNet is depicted in Figure \ref{fig:architecture}.

\begin{figure}[htb]
    \centering
    \includegraphics[width=0.5\linewidth]{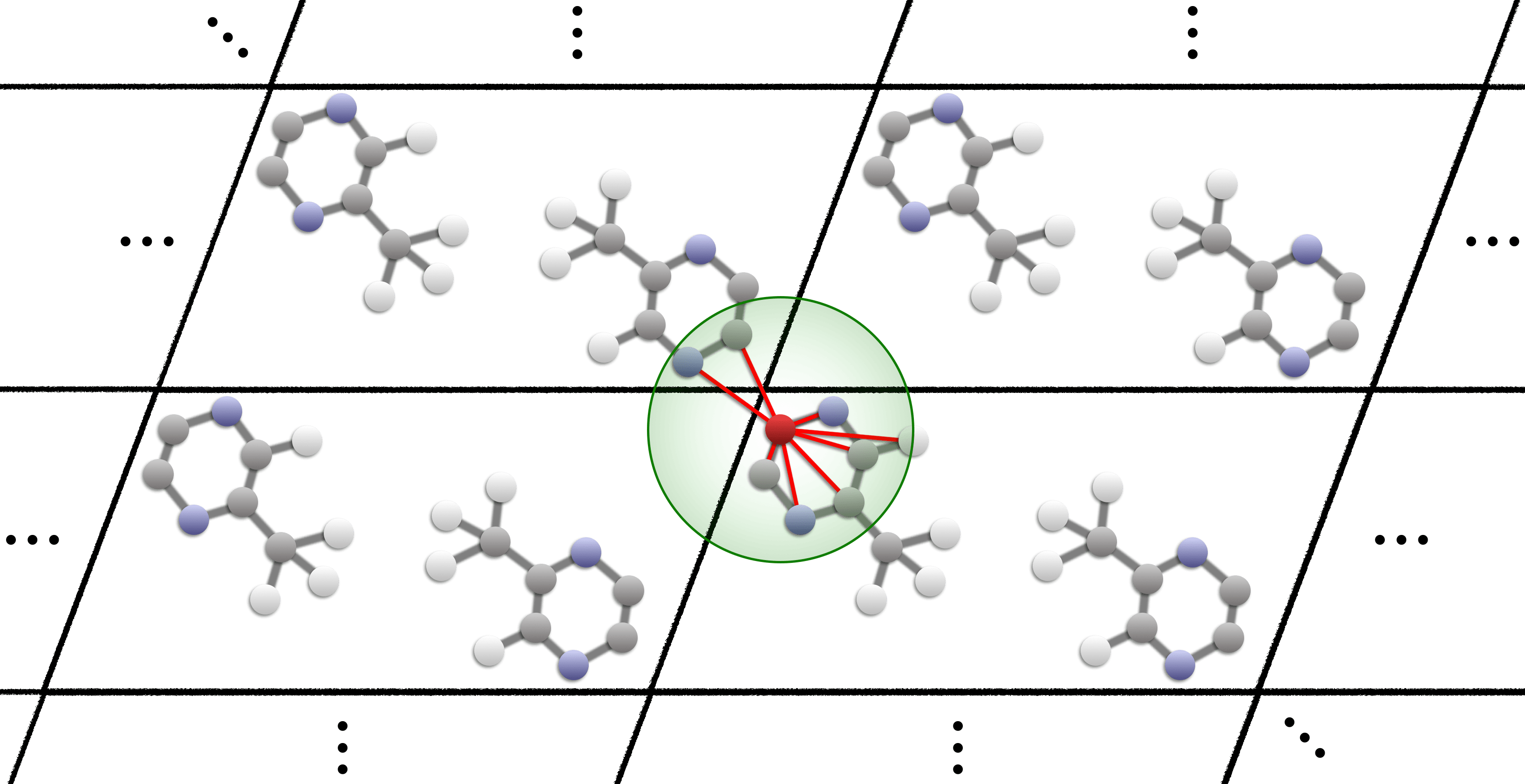}
    \caption{Representation of the graph construction process for the atom highlighted in red colour. Covalent bonds are ignored, and a radius around the atom (depicted in green) is defined. Any atom within this radius is considered a neighbour of the red atom and is connected in the graph (depicted using red lines). Periodic boundary conditions are employed to replicate the infinite nature of the crystal.}
    \label{fig:pbc-graph}
\end{figure}

To construct the graph representation, we employ a radius-based neighbourhood approach using PBCs for all the atoms in the unit cell.
Figure \ref{fig:pbc-graph} shows an example of how the graph is created for a single atom.
Specifically, a cutoff radius ($r_c = 5$~\AA{}) is defined around each central atom, and all atoms within this radius are considered part of the local neighbourhood.
The choice of $5$~\AA{} is based on the analysis of the intermolecular interaction distances described in previous works.~\cite{santiago}.
The graph obtained by this neighbourhood captures both short-range, usually covalent bonds, and relevant weaker intermolecular interactions, such as hydrogen bonds, $\pi$–$\pi$ stacking, halogen bonds, cation–$\pi$ and anion–$\pi$, or van der Waals interactions.
By including all neighbour atoms within this defined radius, the graph effectively represents each atom's local environment, which is crucial for accurately modelling the system's behaviour.

\subsubsection{Atom Encoder}

\begin{figure}[htb]
    \centering
    \includegraphics[width=0.2\linewidth]{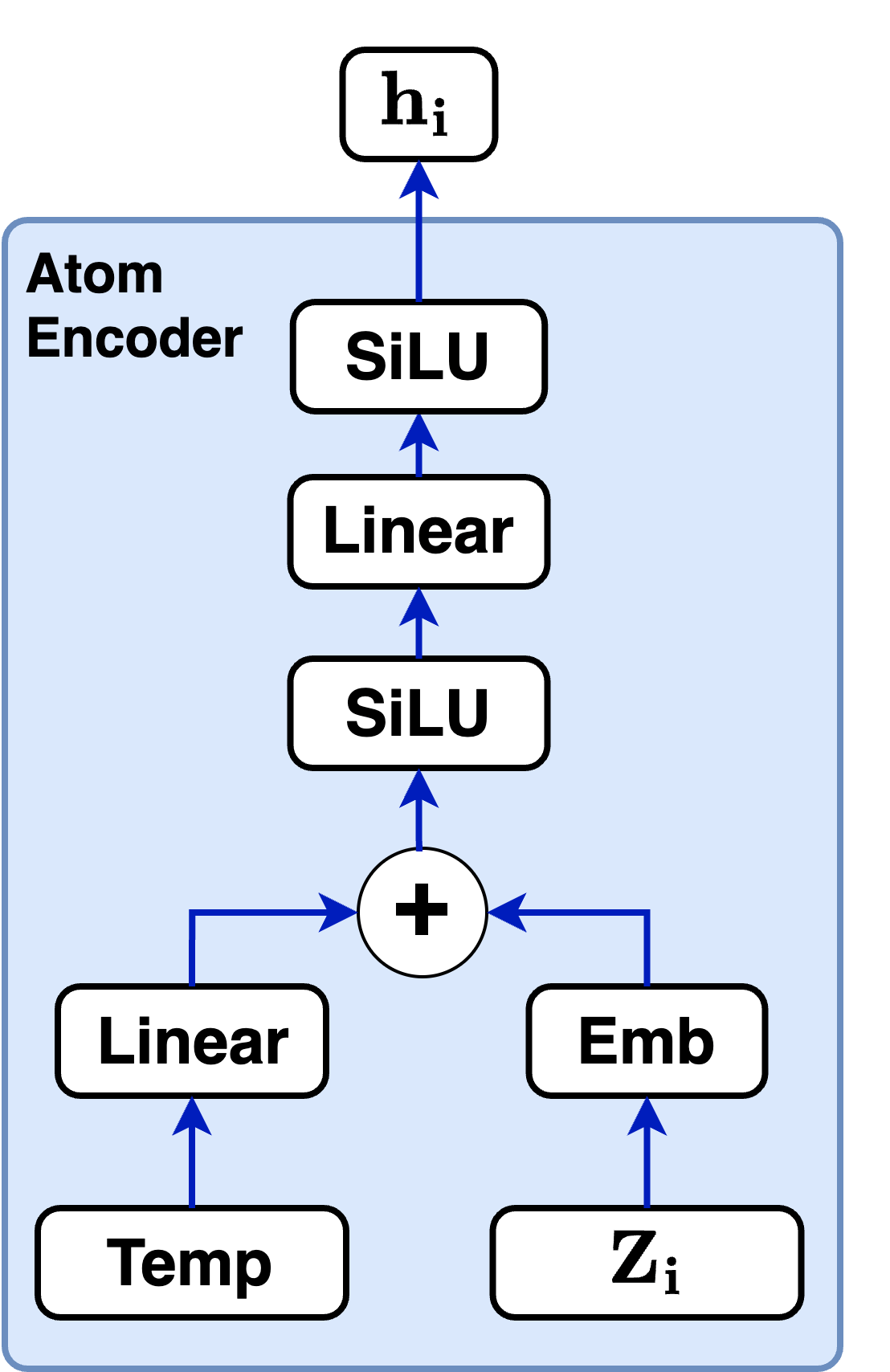}
    \caption{Schematic of the Atom Encoder used in CartNet. The encoder processes each atomic number ($Z_i$) using an embedding layer and its temperature using a linear layer. The resulting encoded features are summed and passed through a $SiLU$ activation function, followed by an additional linear layer and another $SiLU$. The final vector $h_i$ serves as the initial node features for the CartLayer.}
    \label{fig:atom_encoder}
\end{figure}

The atom encoder is responsible for encoding the input information of each atom in the graph. 
In the case of ADPs, this corresponds, for each atom, to its tomic number and the global temperature of the crystal structure. 
Figure \ref{fig:atom_encoder} shows a schematic of the Atom Encoder used in CartNet.

The atom type is encoded using an Embedding~\cite{bengio2000neural} layer, which generates a feature vector corresponding to each distinct atom type.
The temperature is standardised using the training temperature statistics to achieve zero mean and unitary standard deviation. 
The standardised temperature is passed through a Linear layer to ensure dimensional compatibility with the atom-type feature vector.
The resulting temperature and atom-type feature vectors are combined, passed through a $SiLU$~\cite{elfwing2018sigmoid} activation function, and followed by another linear layer and another $SiLU$.
The Equation (\ref{eq:atom_encoder}) describes the encoding of the atom.

\begin{equation}
    \mathbf{h_i} = SiLU(\mathbf{W_2}(SiLU(Emb(Z_i) + \mathbf{W_1}(T) + \mathbf{b_1})) + \mathbf{b_2})  \in \mathbb{R}^{dim}
    \label{eq:atom_encoder}
\end{equation}

Where $Z_i$ represents the atomic number, $Emb$ is an Embedding Layer $\in \mathbb{R}^{2din}$, $T$ is the standardized temperature in Kelvin,  $\mathbf{W_1} \in \mathbb{R}^{1\times 2dim}$, $\mathbf{W_2} \in \mathbb{R}^{2dim\times dim}$, $\mathbf{b_1} \in \mathbb{R}^{2dim}$, $\mathbf{b_2} \in \mathbb{R}^{dim}$, and $dim$ is the number of dimensions of the latent vector.

\subsubsection{Edge Encoder}

\begin{figure}[htb]
    \centering
    \includegraphics[width=0.2\linewidth]{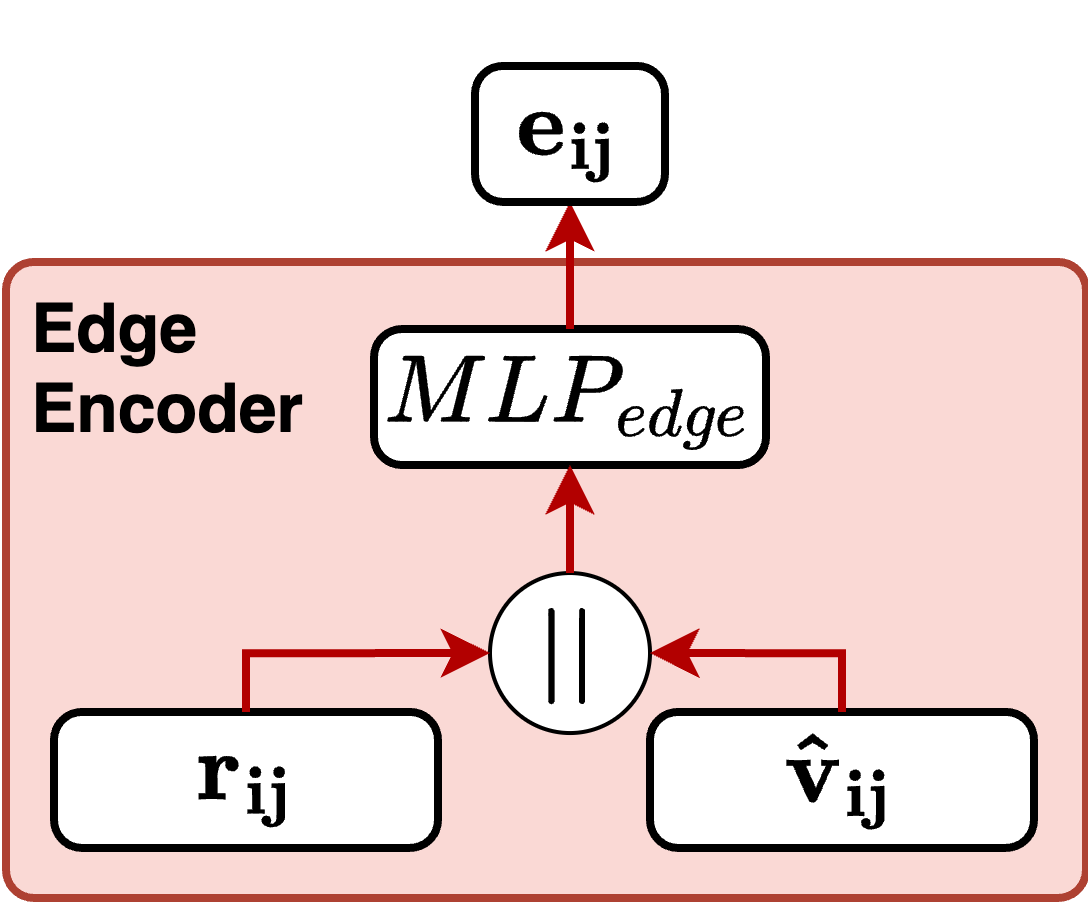}
    \caption{Schematic of the Edge Encoder used in CartNet. The $\mathbf{e_ij}$ and the $\mathbf{\hat{v}_{ij}}$ are concatenated and then processed by a $MLP_{edge}$ to generate the initial edge features utilized by the CartLayer.}
    \label{fig:bond-encoder}
\end{figure}

The edge encoder is responsible for encoding the geometric relationships between atoms in the system.
Figure \ref{fig:bond-encoder} shows a schematic of the Edge Encoder used in CartNet.

The edge is defined as the connection between the receiving atom $i$ and the sender atom $j$.
From each edge, the Euclidean distance ($d_{ij}$) and the direction vector ($v_{ij}$) are calculated based on their positions $\mathbf{p}$.
Equation (\ref{eq:distance}) and (\ref{eq:direction}) define the distance and direction vector, respectively.

\begin{equation}
    d_{ij} =\|\mathbf{p_j}-\mathbf{p_i}\|
    \label{eq:distance}
\end{equation}

\begin{equation}
    \hat{v}_{ij} = \frac{\mathbf{p_j}-\mathbf{p_i}}{d_{ij}}
    \label{eq:direction}
\end{equation}

Where $\mathbf{p_i}$ and $\mathbf{p_j}$ are the receiver and sender atom positions.
The distance is encoded through a Radial Basis Function (RBF) of $K$ elements, transforming the scalar distances into a higher-dimensional space, allowing for a more nuanced representation of geometric relationships, as proposed by previous works in molecular systems~\cite{simeon2024tensornet}.
Equation (\ref{eq:gauss_rbf}) defines the RBF at each $k$ element.

\begin{equation}
    r_k\left(d_{i j}\right)=\exp \left(-\beta\left(\exp \left(-d_{i j}\right)-\mu_k\right)^2\right)
    \label{eq:gauss_rbf}
\end{equation}

Where $\beta$ and $\mu_k$ are fixed values that determine the centre and width of the $k$-th radial basis function, and $K$ is the number of RBF basis. 
The $\mu_k$ values are equally spaced between $\exp(-r_c)$ and 1, while $\beta$ value is equal to $\left[ 2K^{-1}(1 - \exp(-r_c)) \right]^{-2}$ for all $k$.
Here, $r_c$ represents the cutoff radius distance used to define the neighbourhood.

Equation (\ref{eq:rbf}) formalizes the concatenation of all $r_k$ values into vector $\mathbf{r}_{ij}$ to encode distances $d_{ij}$ in a higher dimensional space.  

\begin{equation}
    \mathbf{r_{ij}} = \left[ r_0, r_1,..., r_{K-1}\right]  \in \mathbb{R}^K
    \label{eq:rbf}
\end{equation}

The director vector, $\mathbf{v}_{ij}$, and the RBF-transformed distances, $\mathbf{r}_{ij}$, are concatenated and passed through a Multi-Layer Perceptron ($MLP$) to produce the edge feature vectors.
The $MLP_{edge}$ consists of one first linear layer that doubles the existing dimension, a $SiLU$, another linear layer that returns to the original dimension, and a final $SiLU$. 
This distance and direction information combination ensures that the edge encoding captures the geometric relationships necessary for accurate modelling the geometric structure.
The Equation (\ref{eq:edge_encoder}) describes the edge encoding process mathematically.

\begin{equation}
    \mathbf{e_{ij}} = MLP_{edge}(\mathbf{r_{ij}}(d_{ij})||\mathbf{\hat{v}_{ij}})  \in \mathbb{R}^{dim}
    \label{eq:edge_encoder}
\end{equation}

\subsubsection{CartLayer}

\begin{figure}[htb]
    \centering
    \includegraphics[width=0.3\linewidth]{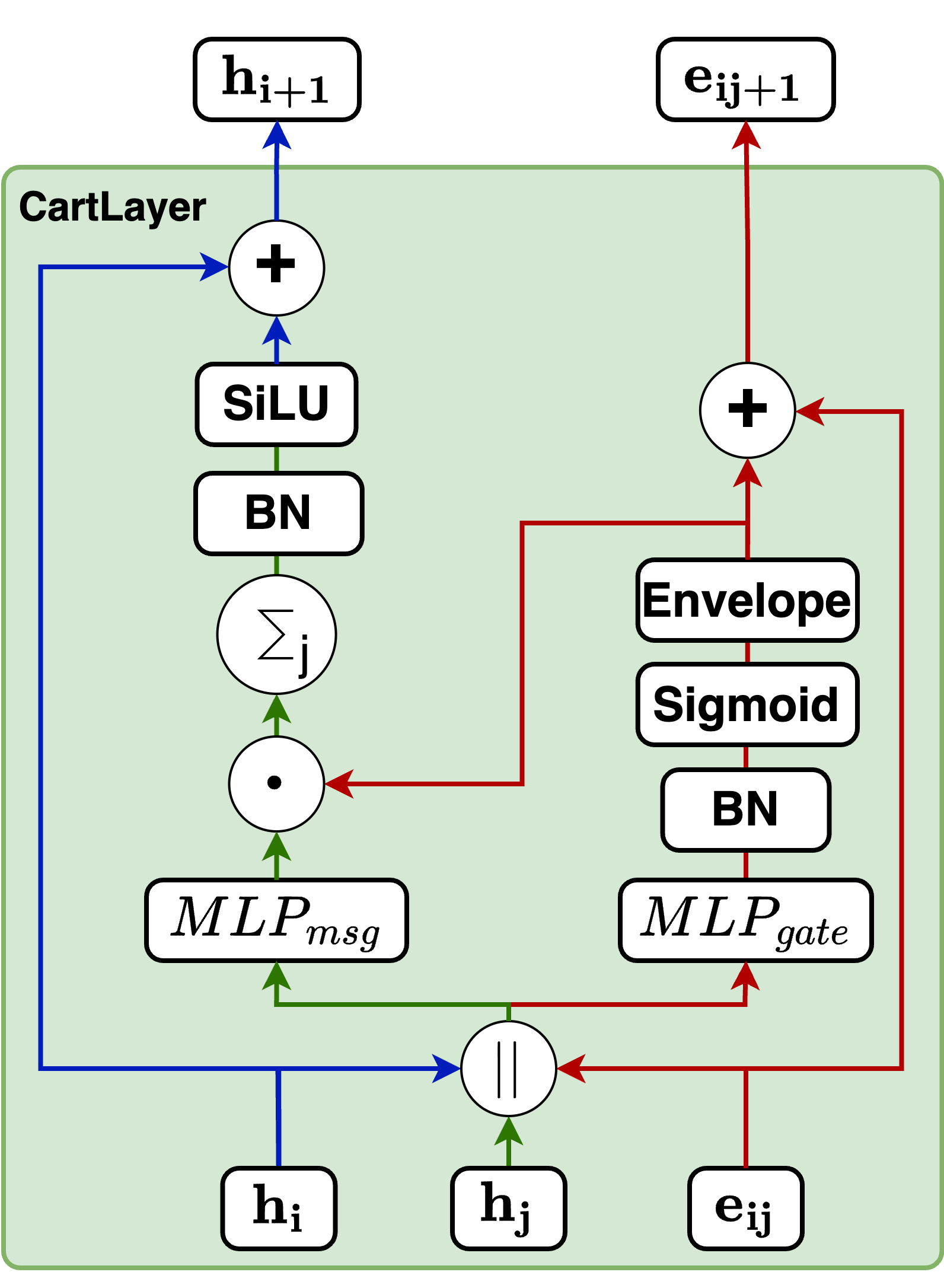}
    \caption{Schematic of the CartLayer used in CartNet. This layer aggregates information between two neighbouring atoms, the sender atom $\mathbf{h}^l_i$ and the receiver atom $\mathbf{h}^l_j$, and their corresponding edge, $\mathbf{e}^l_{ij}$. It utilizes this aggregated information to update the receiving node's features and the edge's latent vector. The updated receiving node, $\mathbf{h}^{l+1}_{i}$, and the respective edge, $\mathbf{e}^{l+1}_{ij}$, representations are then propagated to subsequent layers for further processing.}
    \label{fig:cartnet-layer}
\end{figure}

The CartLayers are responsible for aggregating the information between nodes through message passing. 
It consists of two key components: the gating mechanism and the message-passing mechanism.
Figure \ref{fig:cartnet-layer} illustrates a schematic of the CartLayer.

Our gating mechanism considers the sender and receiver atoms and the edge attributes connecting them and processes them through an $MLP_{gate}$.
The $MLP_{gate}$ consist of a linear layer that reduces the dimensions from $3dim$ to the $dim$, followed by a $SiLU$, and another linear layer that does not modify the dimensions.  
The gating mechanism has a dual purpose: determining the weight of the message and updating the edge attributes.
Additionally, our gating mechanism incorporates an Envelope function inspired by previous work in molecular systems~\cite{simeon2024tensornet}.
Equation (\ref{eq:gate}) describes the gating mechanism.

\begin{equation}
    \mathbf{gate_{ij}} = \text{Sigmoid}(\text{BN}(\text{MLP}_{\text{gate}}(\mathbf{h_i} || \mathbf{e_{ij}} || \mathbf{h_j}))) \odot \text{Env}(d_{ij}) \in \mathbb{R}^{dim}
    \label{eq:gate}
\end{equation}

\begin{equation}
    \text{Env}\left(d_{i j}\right)=\frac{1}{2}\left(\cos \left(\frac{\pi d_{i j}}{r_c}\right)+1\right)
    \label{eq:envelope}
\end{equation}

Here, $\mathbf{h_i}$ and $\mathbf{h_j}$ represent the hidden feature vectors of the receiver and sender atoms, respectively, while $\mathbf{e_{ij}}$ represents the hidden feature vector of the edge. BN stands for Batch Normalization.
As described by Equation (\ref{eq:envelope}), the Envelope function applies a cosine decay over distance.

The Envelope function equalizes the influence of edges based on distance.
This equalization helps the model to detect the peaks from the distribution, making easier to identify the different interatomic interactions from our dataset.

Moreover, the Envelope function also softens the influence of neighbours near the cutoff distance, which is particularly valuable in noisy situations.
It ensures that atoms near the cutoff radius gradually lose influence, preventing them from being considered neighbours based strictly on a hard cutoff radius.
Since our dataset is derived from experimental data, noise is expected, and the Envelope function provides a robust mechanism to handle such noise effectively.
Supplementary Information Section S1 provides further information about the Envelope.

Our message-passing mechanism constructs messages by processing the concatenated sender and receiver atom information and edge attributes through an $MLP_{msg}$.
The $MLP_{msg}$ has the same configuration as the $MLP_{gate}$.
Once the message is created, it is weighted by the gate function, which determines the relative importance of each feature within the message vector.
The weighted messages are aggregated at the receiving node and passed through a Batch Normalization layer, followed by a SiLU non-linearity.
The gate mechanism also updates the edge features, ensuring that the edge attributes remain consistent with the evolving node representations.
To mitigate the issue of gradient vanishing, a skip connection is incorporated into both the node and edge updates, preserving the flow of information through the network layers.
The Equations (\ref{eq:message}), (\ref{eq:node_update}), and (\ref{eq:edge_update}) mathematically describe the message used for message passing and how the atoms and the edges are updated.

\begin{equation}
    \mathbf{msg_{ij}} = \text{MLP}_{\text{msg}}(\mathbf{h_i} || \mathbf{e_{ij}} || \mathbf{h_j}) \odot \mathbf{gate_{ij}}  \in \mathbb{R}^{dim}
    \label{eq:message}
\end{equation}

\begin{equation}
    \mathbf{h^{l+1}_{i}} = \mathbf{h^l_i} + \text{SiLU}(\text{BN}(\sum_{j \in \mathcal{N}_i} \mathbf{msg_{ij}}))  \in \mathbb{R}^{dim}
    \label{eq:node_update}
\end{equation}

\begin{equation}
    \mathbf{e^{l+1}_{ij}}=\mathbf{e^{l}_{ij}}+\mathbf{gate_{ij}}  \in \mathbb{R}^{dim}
    \label{eq:edge_update}
\end{equation}

Here, $\mathbf{h_{i+1}}$ and $\mathbf{e_{ij+1}}$ represent the updated node and edge features, respectively, $\mathbf{msg_{ij}}$ denotes the message vector created between atoms $i$ and $j$, and $\mathcal{N}_i$ is the neighbourhood from the receiving atom.

\subsubsection{Cholesky Head}

The head of our model is designed using Cholesky decomposition to ensure that all output matrices are symmetric and positive-definite, which is a critical requirement for the ADPs.
Figure \ref{fig:cartnet-layer} shows a schematic of the Cholesky Head used in CartNet.

\begin{figure}[htb]
    \centering
    \includegraphics[width=0.2\linewidth]{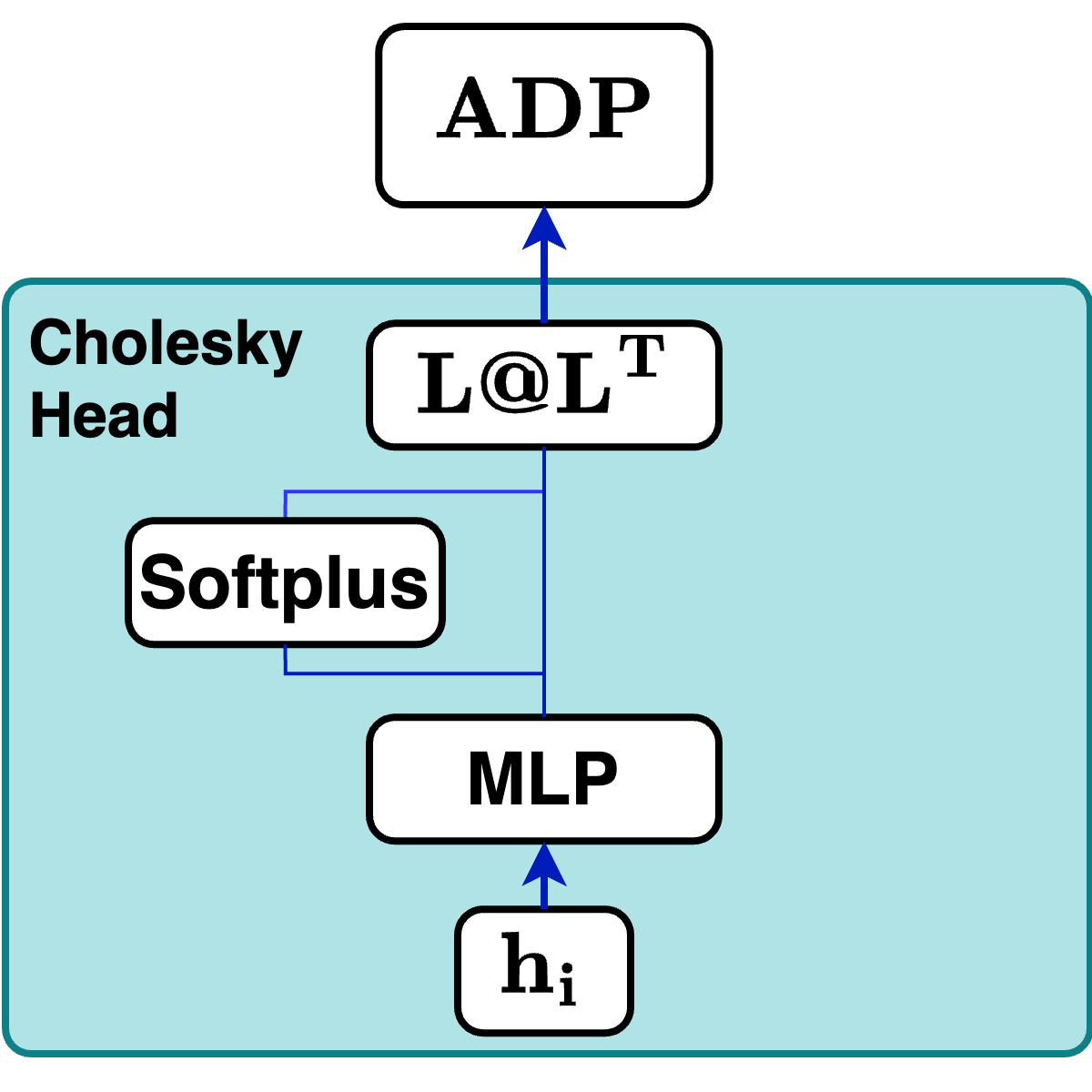}
    \caption{Schematic of the Cholesky Head used in CartNet. This layer enforces the creation of symmetric and positive-definite output matrices, which are necessary conditions for ADP matrices. The final hidden state $ \mathbf{h_i} $ is processed by an $MLP_{head}$ that outputs a vector of six elements. The first three elements are activated using a Softplus function, while the remaining three remain unchanged. These six elements are utilized to construct the lower-triangular matrix $ \mathbf{L} $. The final ADP representation is obtained by multiplying $ \mathbf{L} $ with its transpose, resulting in $ \mathbf{L}\mathbf{L}^\top $.}
    \label{fig:cholesky-head}
\end{figure}

The Cholesky decomposition says that any symmetric positive-definite matrix, such as the ADPs, can be uniquely decomposed into the product of a lower triangular matrix and its transpose. 
This decomposition can be expressed with Equation (\ref{eq:U}).

\begin{equation}
    \mathbf{U} = \mathbf{L}\mathbf{L^T} \in \mathbb{R}^{3\times3}
    \label{eq:U}
\end{equation}

Where $\mathbf{L}$ is a lower triangular matrix, described by Equation (\ref{eq:L_U}).

\begin{equation}
\mathbf{L} = \begin{bmatrix}
l_{11} & 0 & 0 \\
l_{21} & l_{22} & 0 \\
l_{31} & l_{32} & l_{33}
\end{bmatrix} \in \mathbb{R}^{3\times3}
\label{eq:L_U}
\end{equation}

In this matrix, the diagonal elements $l_{11}$, $l_{22}$, and $l_{33}$ are always positive, ensuring that the resulting matrix $\mathbf{U}$ is both symmetric and positive-definite.

Based on this mathematical foundation, the feature vector of each node from the final aggregation layer is processed through a $MLP_{head}$ to produce a feature vector $o_i$ with $i=1,..,6$.
The $MLP_{head}$ consist of a linear layer that reduces the dimensions from $dim$ to $dim/2$, followed by a $SiLU$, and another linear layer that reduces the dimensions from $dim/2$ to $6$.
The first three elements are activated using the $SoftPlus$ function~\cite{nair2010rectified}.
In this context, the $SoftPlus$ activation ensures that the diagonal elements of the matrix $\mathbf{L}$ are strictly positive, which is essential for maintaining the positive-definite property of the output matrix.
The remaining three elements of the feature vector are used as the lower off-diagonal elements of the matrix $\mathbf{L}$.
The construction of the matrix $\mathbf{L}$ is as follows, where the first three elements of the feature vector are placed on the diagonal, and the remaining elements are placed in the lower triangular part, as can be seen in Equation (\ref{eq:L_chol}).

\begin{equation}
\mathbf{L} = \begin{bmatrix}
Softplus(o_1) & 0 & 0 \\
o_4 & Softplus(o_2) & 0 \\
o_6 & o_5 & Softplus(o_3) 
\end{bmatrix} \in \mathbb{R}^{3\times3}
\label{eq:L_chol}
\end{equation}

Finally, the ellipsoid matrix $\mathbf{U^{pred}}$ is obtained by performing a matrix multiplication between $\mathbf{L}$ and its transpose, as can be seen in Equation (\ref{eq:cholesky}).

\begin{equation}
    \mathbf{U^{pred}} = \mathbf{L L^T} \in \mathbb{R}^{3\times3}
    \label{eq:cholesky}
\end{equation}

This construction ensures that the predicted ellipsoid matrix $\mathbf{U_{i}^{pred}}$ is always symmetric and positive-definite, which is essential for accurate modelling the ellipsoid matrices.

\subsection{Rotation SO(3) Augmentation}

Rotation SO(3) augmentation was applied to promote the model to generalize to unseen rotations. 
This augmentation was implemented by multiplying a random three-dimensional rotation matrix with the direction vector between two atoms, as described by the Equation (\ref{eq:rot_vect}).

\begin{equation}
    \mathbf{\hat{v}^{aug}_{ij}} = \mathbf{R}  \mathbf{\hat{v}_{ij}} \in \mathbb{R}^{1\times3}
    \label{eq:rot_vect}
\end{equation}

In this equation, $\mathbf{R}$ is a random rotation matrix, where $\mathbf{R} \in \mathbb{R}^{3\times3}$, and $\mathbf{\hat{v}_{ij}}$ represents the direction vector between two atoms. 
Our approach ensures that the model is regularly exposed to diverse rotational configurations during training, helping it learn features that generalize to previously unseen orientations. 
Although this effectively increases the complexity of the learning problem, online data augmentation is used to keep the additional overhead manageable.
By adopting this well-established deep learning technique, we can rely on simpler model architectures without strictly enforcing equivariance, thereby reducing the computational cost per prediction and enhancing overall efficiency.

The situation is particularly nuanced for the ADP Dataset because ellipsoids are inherently rotationally equivariant. 
Therefore, the ellipsoids must rotate consistently with the input data during augmentation.
The Equation (\ref{eq:rotation_adp}) describes how the rotation is applied to the original $\mathbf{U_i}$ to rotate it.
The proof of Equation (\ref{eq:rotation_adp}) can be found in the Supplementary Information Section S2.

\begin{equation}
    \mathbf{U^{aug}_i} = \mathbf{R} \mathbf{U_i}  \mathbf{R^T} \in \mathbb{R}^{3\times3}
    \label{eq:rotation_adp}
\end{equation}

\section{Experiments and Results}

\subsection{Computational Details}
\label{sec:setup}

The computational setup for all our experiments consisted of an NVIDIA RTX 3090 GPU with 24GB of memory and a system powered by 2x AMD EPYC 7313 16-core CPUs. 
All code implementations used PyTorch v1.13.1~\cite{pytorchwebpage} and PyTorch Geometric v2.3.~\cite{pytorchgeometric}.
Theoretical calculations used Vienna ab initio Simulation (VASP) c6.4.3~\cite{vasp1,vasp2,vasp3} and Phonopy v2.19.1~\cite{phonopy-phono3py-JPSJ, phonopy-phono3py-JPCM}, and were computed on the MareNostrum 5 HPC from the Barcelona Supercomputing Centre (BSC).
The code is publicly available in the Github repository: \href{https://github.com/imatge-upc/CartNet}{https://github.com/imatge-upc/CartNet}.

\subsection{Results}

To evaluate our model, we first tested its prediction performance on two well known public datasets (Jarvis Dataset~\cite{choudhary2020joint} and the Materials Project Dataset~\cite{doi:10.1021/acs.chemmater.9b01294}) and our proposed ADP dataset.
By applying our method to both proprietary and public datasets, we aimed to demonstrate its robustness and generalizability in predicting material properties under various conditions.
It is important to note the very different nature of the materials and properties analysed among the three datasets. 
The two public datasets contain mainly simple bulk materials with properties extracted from electronic structure calculations, and the one developed in this work contains molecular materials using structural information and ADPs from experimental data.

\subsubsection{Jarvis Dataset Results}

The Jarvis 3D DFT Dataset (2021.8.18)~\cite{choudhary2020joint,alignn} is a comprehensive dataset consisting of approximately $55\text{k}$ materials, basically bulk 3D materials, where various DFT properties were computed.
The geometries of the crystal structures were optimized using the OptB88vdW functional (OPT)~\cite{opt88}, which gives accurate lattice parameters. 
The same functional was employed for the calculation of the different properties.
Although in the case of the band gap, to get a better estimation of the value, additionally a small subset was also calculated with the Tran-Blaha modified Becke Johnson potential (MBJ)~\cite{MBJ}.
The dataset provides fundamental material properties: 
(i) formation energy: the energy change when forming a compound from its elements indicating thermodynamic stability, 
(ii) band gap (OPT): the energy difference between valence and conduction bands from standard DFT calculation, 
(iii) total energy: the ground-state energy of the crystal structure, 
(iv) band gap (MBJ): band gap computed using the Tran-Blaha modified Becke Johnson potential for improved accuracy, and 
(v) ehull: the energy above the convex hull, measuring stability against decomposition into other phases.
Notably, the dataset contains only $18\text{k}$ samples for the Band Gap (MBJ) property, making it a low-data scenario.

In this study, we compared the results of our method against other previously reported methods, including Matformer~\cite{matformer}, PotNet~\cite{potnet}, eComFormer~\cite{conformer}, and iComFormer~\cite{conformer}. 
To ensure a consistent and fair evaluation, we followed the methodology proposed by Matformer~\cite{matformer} and used their proposed data splits. 
We use the mean absolute error (MAE) as our evaluation metric and report its mean and standard deviation across four random initialization seeds to confirm that our model’s performance is robust rather than dependent on initialization.
Section S4 from the Supplementary Information provides the detailed CartNet modifications and training configurations used for predicting each property.

\begin{table}[htb]
\centering

\caption{MAE results for the different tested architectures in test split from the Jarvis dataset. Best result in $\mathbf{bold}$ and second best {\ul underlined}. Arrows indicate the direction of improvement for each metric.}
\label{tab:results-jarvis}
\resizebox{\textwidth}{!}{%
\begin{tabular}{@{}cccccc@{}}
\toprule
Method & Form. Energy (meV/atom)↓ & Band Gap(OPT) (meV)↓ & Total energy (meV/atom)↓ & Band Gap(MBJ) (meV)↓ & Ehull (meV)↓ \\ \midrule
Matformer~\cite{matformer}  & 32.5           & 137           & 35             & 300           & 64             \\
PotNet~\cite{potnet}    & 29.4           & 127           & 32             & 270          & 55             \\
eComFormer~\cite{conformer} & 28.4         &  124           & 32             &  280    &  {\ul 44}       \\
iComFormer~\cite{conformer} &  {\ul 27.2}     & {\ul 122}    & {\ul 28.8}     & {\ul260} &  47            \\
CartNet       &  $ \mathbf{27.05 \pm 0.07} $ &  $\mathbf{115.31 \pm 3.36}$ & $\mathbf{26.58 \pm 0.28}$ & $\mathbf{253.03 \pm 5.20}$         &  $\mathbf{43.90 \pm 0.36}$ \\ \bottomrule
\end{tabular}%
}
\end{table}

As illustrated in Table \ref{tab:results-jarvis}, our model consistently performs best across all evaluated properties. 
The improvements are most notable in the Total Energy and Band Gap (OPT)  predictions, where our model demonstrates an improvement of $7.71\%$ and $5.48\%$, respectively, over the next-best models. 
The low-data Band Gap (MBJ) scenario outperforms the next-best model by approximately $2.68\%$.

Our model's ability to excel across various properties, including those with fewer data samples, such as Band Gap (MBJ), highlights its adaptability and robustness. 
These results suggest that our approach performs well in traditional tasks and thrives in challenging low-data environments, providing a comprehensive solution for crystal structure property prediction.

\subsubsection{The Materials Project Dataset Results}

The Materials Project Dataset-2018.6.1~\cite{doi:10.1021/acs.chemmater.9b01294} contains approximately $69\text{k}$ structures collected from The Materials Project~\cite{materials_project}.
The dataset consists of inorganic crystalline materials, primarily bulk materials, where various DFT properties have been computed.
The structures were optimized using PBE~\cite{PBE}-D3(BJ)~\cite{d3} level of theory.
The dataset provides several essential material properties:
(i) formation energy,
(ii)  band gap, 
(iii) bulk moduli: measures the resistance of a material to deformation under shear stress, reflecting how it deforms when forces are applied parallel to a surface.
(iv) shear moduli: quantifies the resistance of a material to uniform compression, indicating how much it compresses under external pressure.
Notably, the dataset contains only around 5,5k Shear and Bulk Moduli samples, making these properties particularly challenging.
We directly compared our results with those of previous works~\cite{conformer,matformer,potnet} without retraining these models, maintaining consistency across evaluations and using the same data splits. 
Similarly to the previous experiment, we employed the MAE as the evaluation metric with the splits defined in Matformer~\cite{matformer} and ran experiments using four different random seeds, reporting both the mean and standard deviation of the MAE.
The CartNet training configurations for each property are detailed in the Section S4 from the Supplementary Information.

As shown in Table \ref{tab:results-megnet}, our method achieves the best performance across all evaluated properties. 
The improvements are particularly notable for Form. Energy, yielding an approximately $4.33\%$ improvement. 
Similarly, for Bulk Moduli's low-data scenario, our model improves the MAE by approximately $13.16\%$ over the next-best model. 
In the Shear Moduli task, another low-data scenario, our method achieves the same metric as the best know reported.

These results underscore the robustness of our model, especially in low-data environments like the Bulk and Shear Moduli tasks, where limited training data poses significant challenges.
The consistent improvements across all properties confirm that our approach is well-suited for predicting a broad range of crystal structure properties and offers substantial gains over existing methods.

\begin{table}[htb]
\centering
\caption{MAE results for the different tested architectures in test split from the Material Project Dataset. Best result in $\mathbf{bold}$ and second best {\ul underlined}. Arrows indicate the direction of improvement for each metric.}
\label{tab:results-megnet}
\resizebox{\textwidth}{!}{%
\begin{tabular}{@{}ccccc@{}}
\toprule
\multicolumn{1}{c}{Method} &
  \multicolumn{1}{c}{Form. Energy (meV/atom)↓} &
  \multicolumn{1}{c}{Band Gap (meV)↓} &
  \multicolumn{1}{c}{Bulk Moduli (log(GPa))↓} &
  \multicolumn{1}{c}{Shear Moduli (log(GPa))↓} \\ \midrule
Matformer~\cite{matformer}  & 21             & 211          & 0.043  & 0.073  \\
PotNet~\cite{potnet}    & 18,8           & 204          & 0.04   & {\ul 0.065}  \\
eComFormer~\cite{conformer} & {\ul18.16} & 202          & 0.0417 & 0.0729 \\
iComFormer~\cite{conformer} &  18.26    & {\ul193} & {\ul0.038} & \textbf{0.0637} \\
CartNet       & $\mathbf{17.47 \pm 0.38}$          &  $\mathbf{190.79 \pm 3.14}$    & $\mathbf{0.033 \pm 0.94 \cdot 10^{-3}}$     &  $\mathbf{0.0637 \pm 0.0008}$ \\ \bottomrule
\end{tabular}%
}
\end{table}

\subsubsection{ADP Dataset}
\label{sec:res_adp}

Our method has been compared against two other previously reported methods for the ADP Dataset for material property prediction.
eComformer and iComformer~\cite{conformer} have been selected for the comparative since the other methods are based only on distance encoding and do not encode the geometry referenced to a 3D basis needed for the correct prediction of the ADPs direction.
To evaluate the ADPs we computed the MAE between the $U$ matrices.
Also, the Similarity Index ($S_{12}$) was also calculated since is widely used in previous works~\cite{similarity_index} to compare ADPs.
$S_{12}$ is based on the Bhattacharyya distance and represents the percentage error of the overlap between two multivariate Gaussian distributions~\cite{similarity_index}.

As shown in Figure~\ref{fig:uequi}, small ADPs differ by several orders of magnitude compared to larger ones. 
This disparity could lead to biased conclusions when analysing the results using MAE, although the $S_{12}$ does not have this problem. 
Nevertheless, the issue with the $S_{12}$ is that it is not highly discriminative. 
Most of the errors are less than 1\% using this metric, which might give the impression that the predictions are accurate.
To address these issues, we also employed an Intersection over Union (IoU) metric over the ADPS graphical ellipsoid representations. 
The IoU metric provides a measurement independent of the ADP’s size, enabling us to assess whether smaller or larger ellipsoids are well-predicted. 
Furthermore, if an ADP is not well-predicted, it will have a higher impact on the IoU metric, making it more restrictive. 
The IoU is computed by voxelizing the 3D space. 
More details about the IoU implementation and the training configurations can be found in the Section S3 and S4 from the Supplementary Information, respectively.
For all models, we ran experiments using four initialization seeds to ensure robustness against initialization effects, reporting the mean and standard deviation.

Table \ref{tab:results_ellipsoid} shows that our method achieves the best performance in all evaluation metrics, MAE, $S_{12}$ and IoU, compared with other methods. 
In our experiments, CartNet outperforms the second-best model by $10.87\%$ in MAE, $17.58\%$ in $S_{12}$, and $2\%$ in IoU.
Furthermore, CartNet needs to train $49\%$ fewer parameters than the second-best model.
This result suggests that our approach can achieve state-of-the-art results without needing specific layers to enforce the rotational equivariance.
The error introduced by using our data augmentation instead of using equivariant layers is discussed and evaluated in Section \ref{sec:data-aug}.

\begin{table}[ht]
\centering
\caption{Results for the different tested architectures in the test split from the ADP Dataset. Best result in $\mathbf{bold}$ and second best {\ul underlined}. Arrows indicate the direction of improvement for each metric.}
\label{tab:results_ellipsoid}
\begin{tabular}{@{}ccccc@{}}
\toprule
Method     &  MAE ($\text{\AA}^2$)↓ & $S_{12}$ (\%)↓  & IoU (\%)↑   &    \#Params↓   \\ \midrule
eComformer~\cite{conformer} &  $6.22 \cdot 10^{-3} \pm 0.01 \cdot 10^{-3}$ & $2.46 \pm 0.01$ & $74.22 \pm 0.06$ &  5.55M \\
iComformer~\cite{conformer} &  {\ul $3.22 \cdot 10^{-3} \pm 0.02 \cdot 10^{-3}$} &  {\ul $0.91 \pm 0.01$}  &  {\ul $81.92 \pm 0.18$}  &   {\ul 4.9M} \\
CartNet       & $\mathbf{2.87 \cdot 10^{-3} \pm 0.01 \cdot 10^{-3}}$  & $\mathbf{0.75 \pm 0.01}$   & $\mathbf{83.56 \pm 0.01}$   &  \textbf{2.5M} \\ \bottomrule
\end{tabular}%

\end{table}

\begin{figure}[ht]
    \centering
    \includegraphics[width=0.5\linewidth]{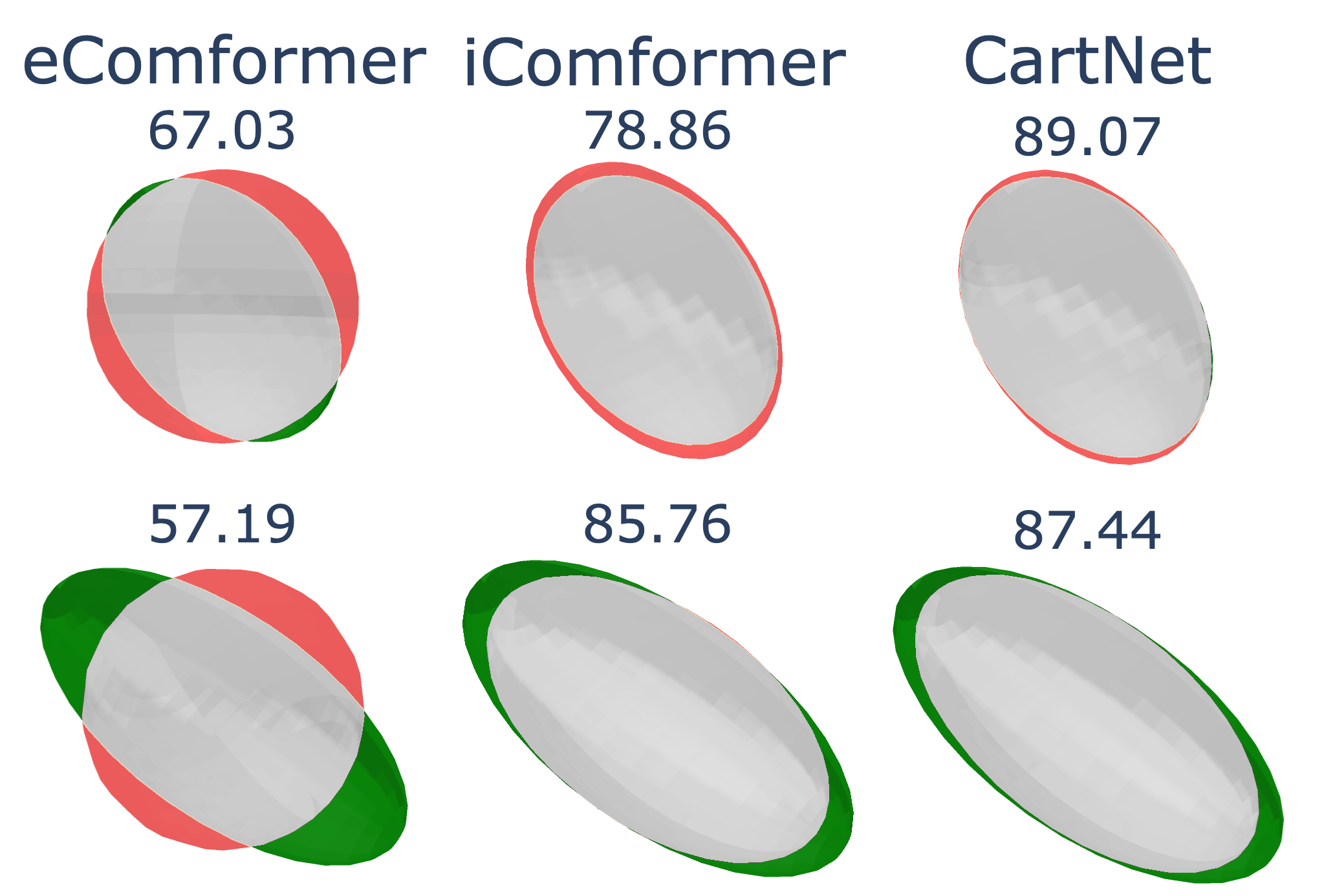}
    \caption{Visual comparison of eComformer, iComformer, and CartNet on the ADP test split. The top row shows an ellipsoid with average anisotropy, the bottom row shows one with high anisotropy. Green indicates experimental values, red shows predicted values, and grey is their intersection. The IoU for each ellipsoid is shown above it.}
    \label{fig:visual-ellipsoids}
\end{figure}

Figure \ref{fig:visual-ellipsoids} shows a visual comparison between the compared methods.
Even though all tested models can encode the 3D geometry, only our approach and iComformer can encode the geometry so that the ellipsoids can be oriented.
On the other hand, eComformer only creates spherical ellipsoids.
If we take a closer look at the eComformer architecture, they create an invariant descriptor based on distance. 
This descriptor is then updated by an equivariant layer based on spherical harmonics and then by a few invariant layers based on distance.
Our intuition suggests that even though this equivariant layer enriches the invariant information and can improve the invariant predictions, it is not able to successfully encode the needed information for a correct ADPs orientation prediction.
Entire crystal structures comparison between methods can be found in the Section S8 from the Supplementary Information.

\begin{figure}[htb]
    \centering
    \includegraphics[width=0.5\linewidth]{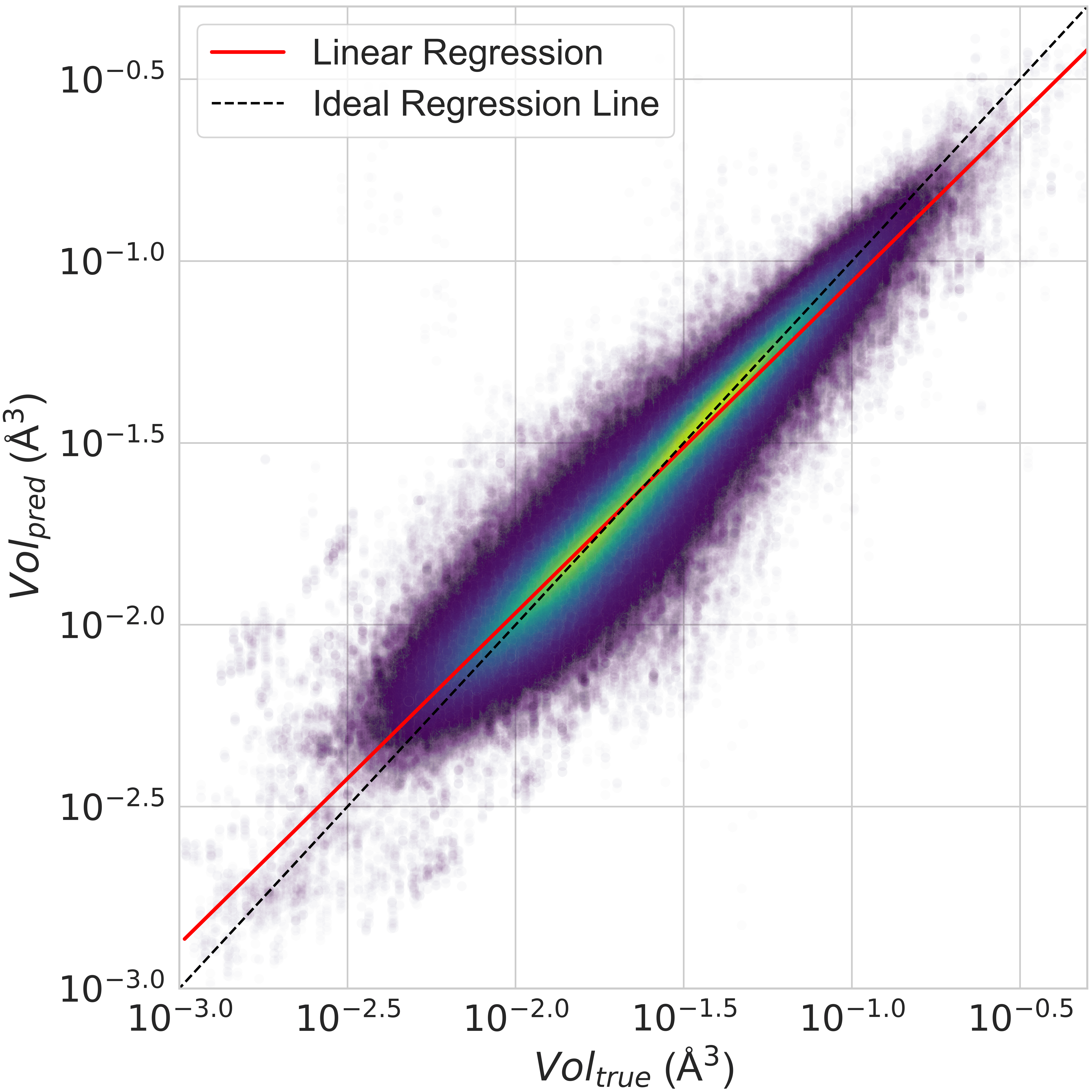}
    \caption{Scatter plot of true versus predicted ADP volumes for the test split of the ADP dataset using CartNet. A linear regression fitted to the data yields  $y = 1.0013x + 0.0021$  with an  $R^2$  score of $0.91$, indicating a correlation. The scale is logarithmic for both axes.}
    \label{fig:regplot}
\end{figure}

Figure~\ref{fig:regplot} presents a scatter plot of the predicted ADP volumes versus the experimental values.
The model achieves a coefficient of determination ($R^2$) of $0.91$, demonstrating a strong linear correlation between the predictions and the actual volumes.
This near-perfect linear regression indicates the model's high accuracy in predicting ADP volumes.

\begin{figure}[htb]
\centering
\begin{minipage}{.45\textwidth}
  \centering
  \includegraphics[width=0.9\linewidth]{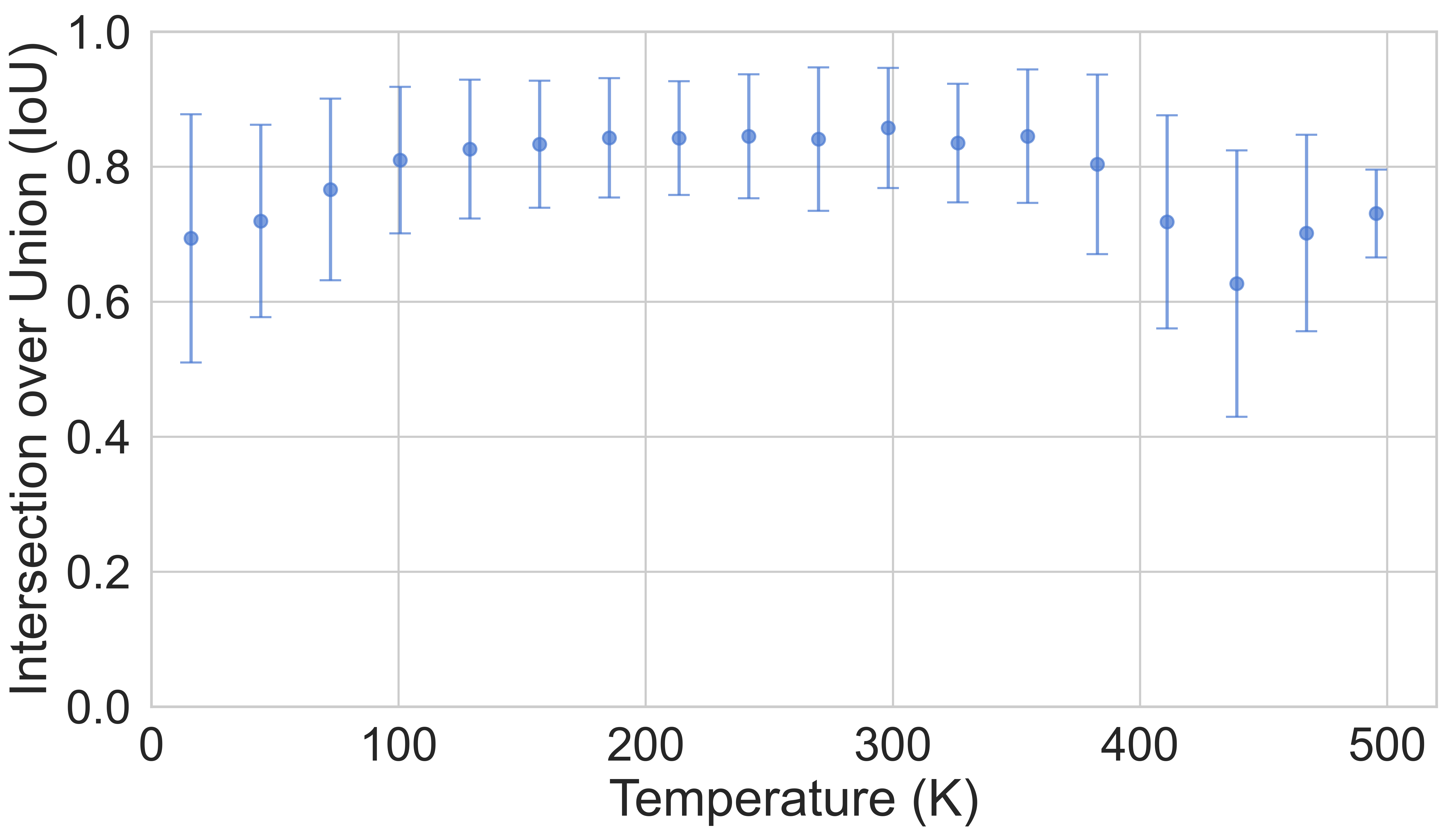}
  \captionof{figure}{Plot of the IoU error with standard deviation as a function of temperature for CartNet’s predictions on the ADP test dataset.}
  \label{fig:error_temp}
\end{minipage}%
\hfill
\begin{minipage}{0.45\textwidth}
  \centering
  \includegraphics[width=0.9\linewidth]{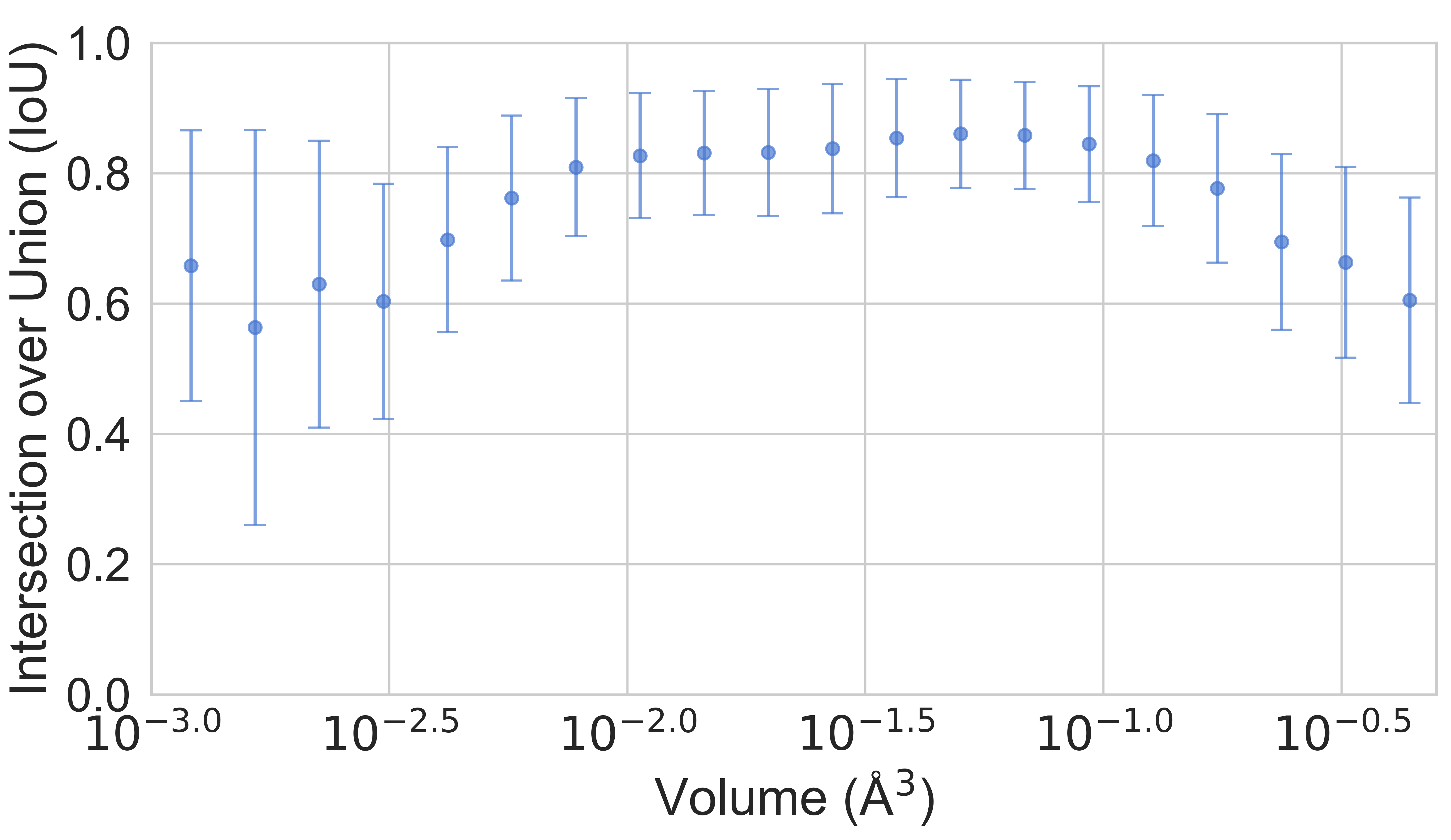}
  \captionof{figure}{Plot of the IoU error with standard deviation as a function of the ADP volume for CartNet’s predictions on the on the ADP test dataset.}
  \label{fig:error_volume}
\end{minipage}
    \centering
    \includegraphics[width=0.6\linewidth]{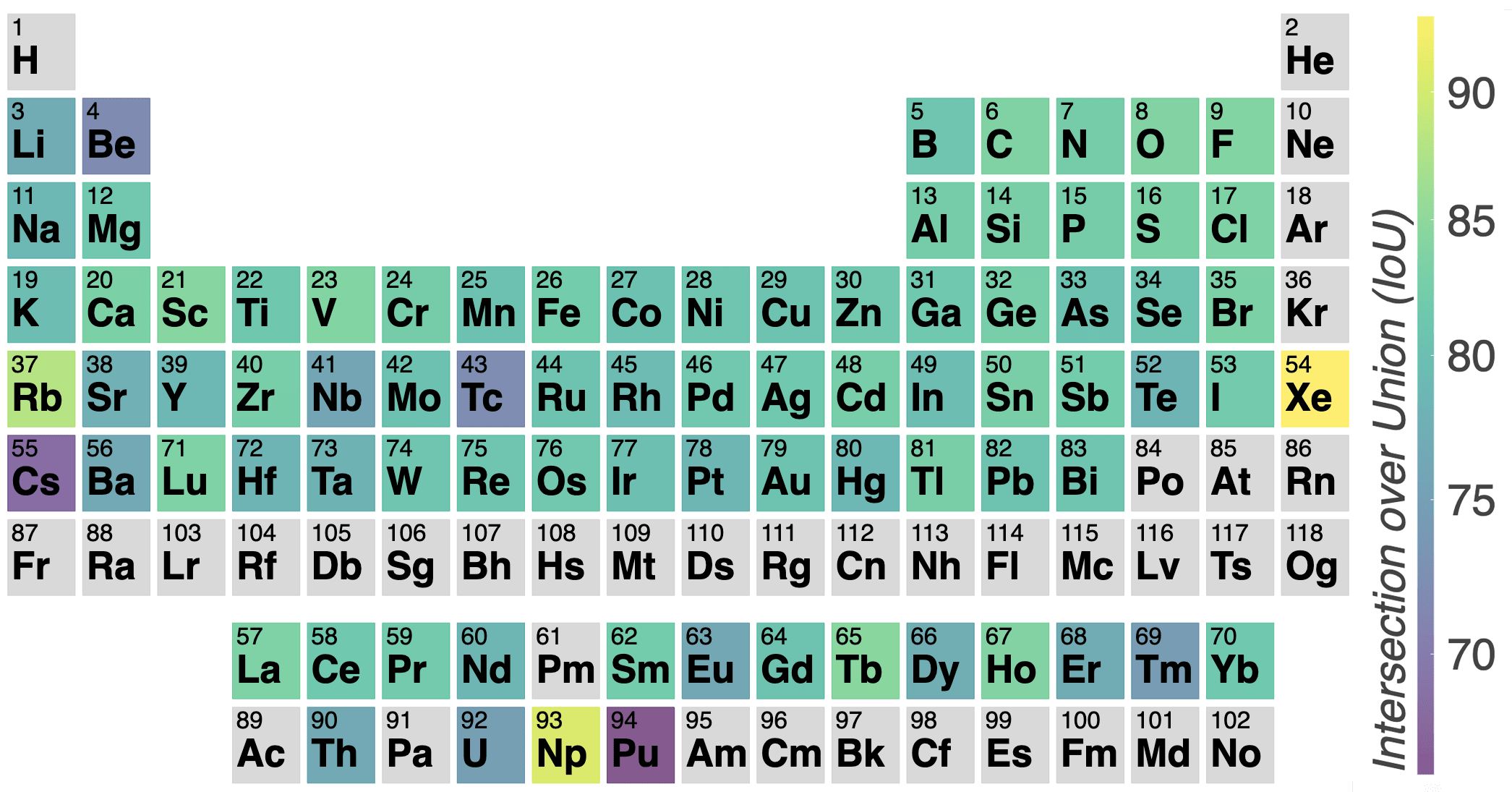}
    \caption{Heatmap illustrating the IoU metric per element for the CartNet's predictions in the test split from the ADP dataset. Lighter colours represent higher IoU, while darker colours indicate lower IoU.}
    \label{fig:error_element}
\end{figure}

In Section~\ref{sec:dataset}, we discussed the data imbalance concerning temperature, volume, and elemental composition in the ADP dataset.
We analysed the error distribution for temperature, volume, and element to determine if the model's errors align with these imbalances.
Figure~\ref{fig:error_temp} shows CartNet's predicted IoU as a function of temperature for the test split of the ADP dataset.
The IoU decreases noticeably when the temperature drops below $80\,\text{K}$ and when it rises above $400\,\text{K}$.
This pattern corresponds with the data distribution depicted in Figure~\ref{fig:hist_temp}, indicating higher errors in temperature regions that are underrepresented in the dataset.
Nevertheless, the IoU is stable for the rest of the temperatures.
Figure~\ref{fig:error_volume} illustrates the IoU metric for CartNet's predictions across different volume ranges in the test split.
Comparing this with Figure~\ref{fig:uequi}, we observe that higher errors occur in volume ranges with fewer data points, especially at the extremes of the chart.
Figure~\ref{fig:error_element} presents the IoU per element for CartNet's predictions on the ADP dataset.
Lower metrics correspond to lower representation in the dataset for some elements, such as Beryllium, Technetium, Caesium, and Plutonium.
Conversely, elements like Xenon and Neptunium exhibit near-perfect IoU scores despite being underrepresented.
Additionally, most of the other elements show similar IoU values, suggesting that the network is able to learn from atom numbers that appear more frequently in the dataset and generalize that information to other atomic numbers.
Further discussion on the performance of different atom types in specific chemical interactions such as hydrogen bonding, $\pi-\pi$ stacking, and tert-butyl groups, as well as additional results on different polymorphs, can be found in the Supplementary Information Section S5.

Moreover, we aimed to investigate whether our model could predict ADPs at various temperatures while using the same crystal geometry.
To this end, we selected a crystal not included in our dataset, which had been synthesised at different temperatures.
The supplementary information from previous works~\cite{thermal_expand} indicated that the series of crystal structures of \textit{guanidinium pyridinium naphthalene-1,5-disulfonate}, with the CSD refcode DOWVOC~\cite{doi:10.1021/cg5014895}, met these criteria.
The CSD contains 14 entries for this crystal, covering a temperature range between $155\,\text{K}$ and $283\,\text{K}$, all with well-defined ellipsoids.
The list of CSD refcodes and the respective temperature can be found in Section S7 from the Supplementary Information.
We conducted two experiments to assess the ability of our model to predict ADPs at any temperature fixing the geometry.
In the first experiment, we examined whether our model could accurately predict the ADPs for all data points of this crystal using the experimental geometries and temperatures.
In the second experiment, we evaluated whether our system could predict ADPs by employing a fixed geometry while varying the input temperature.
We used the fixed geometry at $213\,\text{K}$ and systematically adjusted the input temperature values provided to CartNet.
We computed the mean of the ellipsoid volumes from experimental, predicted, and predicted from the geometry for each temperature point.

\begin{figure}[htb]
    \centering
    \includegraphics[width=0.5\linewidth]{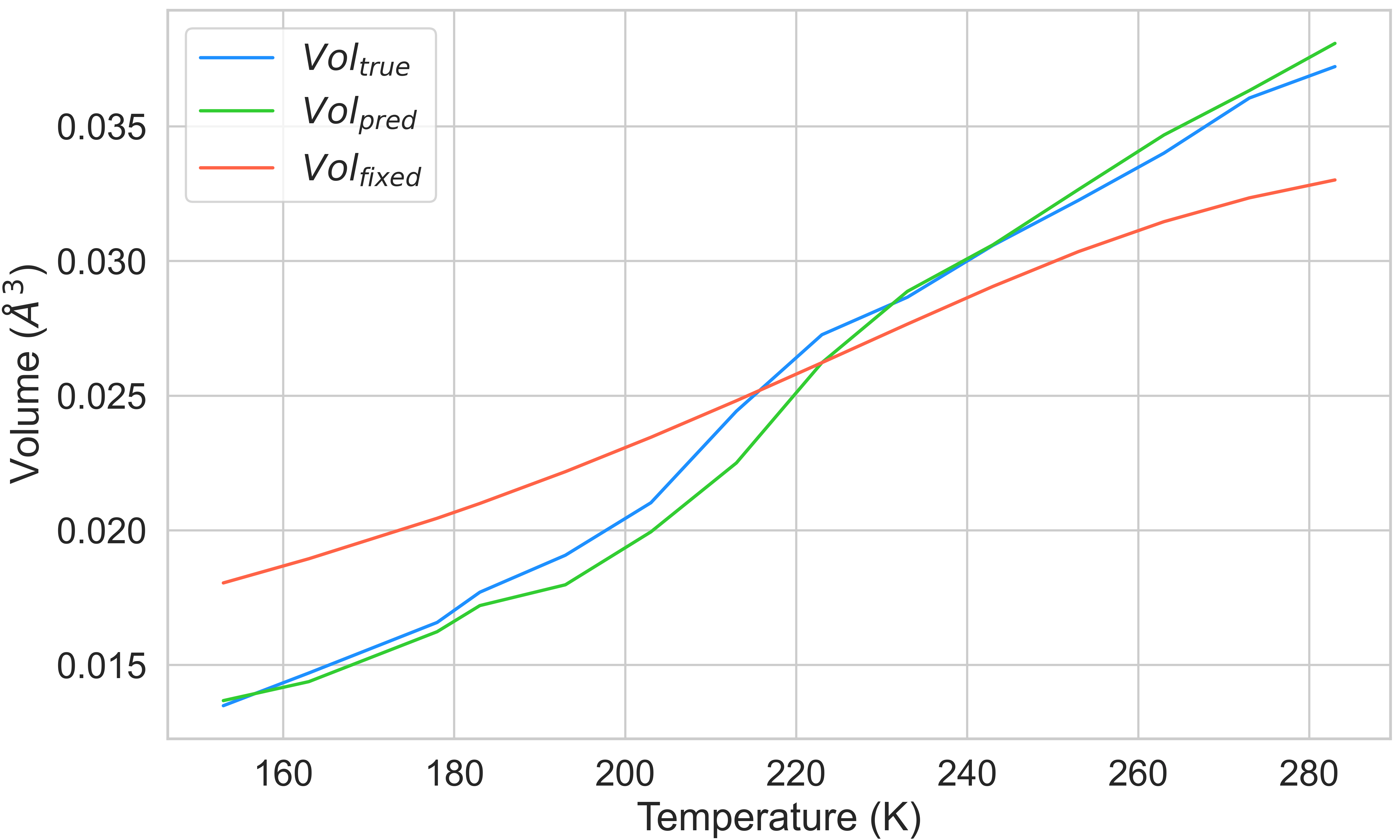}
    \caption{Comparison of the mean ellipsoid volume as a function of temperature for the \textit{guanidinium pyridinium naphthalene-1,5-disulfonate} (CSD refcode: DOWVOC) crystal structure. The blue line shows experimental data, the green line shows predicted volumes using experimental geometry and temperature, and the orange line shows predictions with a fixed 215 K geometry and varying input temperature in CartNet.}
    \label{fig:anchor_middle}
\end{figure}

Figure~\ref{fig:anchor_middle} presents the results of these experiments.
The results demonstrate that our model can predict ADPs across the entire temperature range.
However, the predicted volume diverges when using a fixed geometry and varying the input temperature.
These results suggest that cell expansion due to temperature significantly affects ADPs, as the error increases with the difference between the temperature at which we fixed the geometry and the temperature at which ADPs wanted to be predicted.
Nonetheless, when predicting ADPs around the temperature from the fixed geometry, they are estimated with high accuracy.
Further discussion about using the fixed geometry at 150K and 283K can be found at Section S7 from the Supplementary Information.

\begin{figure}[htb]
    \centering
    \includegraphics[width=\textwidth]{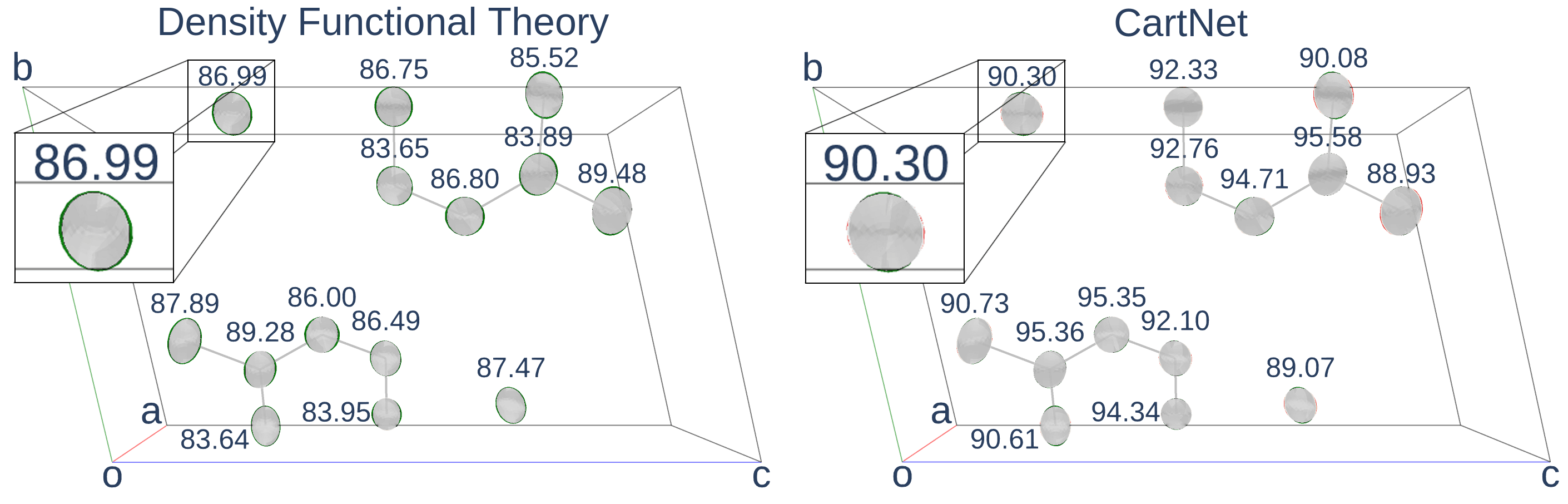}
    \caption{Thermal ellipsoids representations from experimental ADPs for the \textit{5,5'-dimethyl-2,2'-bipyrazine} crystal structure (CSD refcode: ETIDEQ) predicted using DFT and CartNet, respectively. The green regions represent the experimental values, the red ones represent the prediction values, and the grey represents the intersection between them. The numbers in each atom represent the IoU between the experimental and the calculated ADP.
    Highlighted can be seen a sample ellipsoid predicted using the DFT and the same ellipsoid using CartNet. 
    The parallelepiped represents the unit cell, and the red, green, and blue lines correspond to the a, b, and c unit cell axes. }
    \label{fig:phonpy-cartnet}
\end{figure}

Finally, the last experiment was to compare with traditional theoretical methods.
We compared the ADPs of our method with DFT calculations.
Due to the high computational cost of computing ADPs with DFT, a single crystal structure (\textit{5,5'-dimethyl-2,2'-bipyrazine}, CSD refcode: ETIDEQ) has been computed.
The electronic structure calculation was performed using the Vienna ab initio Simulation (VASP c6.4.3)~\cite{vasp1,vasp2,vasp3} package with the optimized structure at PBE~\cite{PBE}-D3(BJ)~\cite{d3} level of theory.
More details about the configuration used in electronic structure calculations can be seen in Section S6 from the Supplementary Information. 

In the case of DFT calculations, the choice of the central geometry for the atomic displacements can have a strong impact on the accuracy of the calculated ADPs. 
In this case, three geometries were tested: 
(i) a full optimization, considering atomic positions and lattice parameters, 
(ii) geometry relaxation including atoms but fixing the lattice to its crystallographic dimensions and 
(iii) an atomic relaxation with a fixed volume, obtained by solving the Vinet equation of state. 
The latter calculation is the most sophisticated since it involves the calculation of the change of free energy with respect to the compression and expansion of the cell to derive a cell volume at a given temperature. 
However, the best results were obtained for the full optimization, which is a simpler method in comparison. 
Regarding the computational cost of this calculation, it is interesting to observe how the Fixed Lattice calculation required half the geometrical displacements compared to Vinet and Full Optimization, as this geometry retained inversion symmetry after structural optimization. 
Thus, the displacements over symmetry equivalent atoms were redundant and, hence, omitted.

Table \ref{tab:phonopy-results} shows the numerical results for this comparison.
As can be seen, our model still improves the MAE by 34.77\% and the IoU by 6.04\%.
Additionally, where CartNet shows real improvement is in the computation time, which was reduced by several orders of magnitude. 
Figure \ref{fig:phonpy-cartnet} compares the ADPs from CartNet and the best DFT results (Full Optimization). In both cases, ellipsoids closely match the experimental reference, in line with their low MAE values.

\begin{table}[ht]
\centering
\caption{Comparative ADP results between CartNet and DFT for the \textit{5,5'-dimethyl-2,2'-bipyrazine} crystal structure (CSD refcode: ETIDEQ). 
For the DFT calculations, three configurations have been tested.
First, using atomic relaxation with a fixed volume, obtained by solving the Vinet equation.
Second, atomic relaxation with a fixed lattice.
Third, full optimization of the geometry.
DFT calculations were done using 56 CPU cores from the MareNostrum 5~\cite{bscMareNostrum} HPC, while CartNet calculations were done using 1 GPU and 1 CPU core from our setup described in Section \ref{sec:setup}. Best result in $\mathbf{bold}$. Arrows indicate the direction of improvement for each metric.}
\label{tab:phonopy-results}
\begin{tabular}{@{}ccccc@{}}
\toprule
Method  & MAE ($\text{\AA}^2$)↓   & $S_{12}$ (\%)↓       & IoU (\%)↑ &    Time (s)↓  \\ \midrule
DFT (Vinet) & $1.32 \cdot 10^{-2}$  &  3.09        & 57.33    & $\sim 2.88 \cdot 10^6 $              \\
DFT (Fix Latt.) & $1.43 \cdot 10^{-2}$  &  4.12        & 70.75      & $\sim 1.44 \cdot 10^6 $                 \\
DFT (Full Opt.) & $3.25 \cdot 10^{-3}$ & 0.49  & 86.27             & $\sim  2.88 \cdot 10^6$                \\
CartNet & $\mathbf{2.12 \cdot 10^{-3}}$  & \textbf{0.17} & \textbf{92.31}     &  $ \mathbf{\sim 10^{-2} }$              \\ \bottomrule
\end{tabular}

\end{table}

\section{Ablation Studies}

We conduct comprehensive ablation studies to assess the contribution of each component in our proposed model. 
By systematically removing or altering specific elements of the architecture, we aim to understand the impact of each part on the overall performance. 
This analysis allows us to identify which components are crucial for achieving high accuracy and provides insights into the model’s inner workings.

Table \ref{tab:ablations} presents the results of removing each component of our proposed method.
The following subsections explain the detailed experiments done and discuss their implications in the design of our model.

\begin{table}[htb]
    \centering
    \caption{Ablation results in the test split of the ADP dataset. Exp. Nº 1 involves the full CartNEt using all contributions, Exp. Nº 2 creates the graph without the hydrogens, Exp. Nº 3 was trained without using the envelope to equalize the neighbours, Exp. Nº 4 was trained without using the direction unit vector between the neighbours, Exp. Nº 5 was trained without using the temperature of the crystal structure as input, and Exp. Nº 6 was trained without the SO(3) data augmentation proposed method. Best result in $\mathbf{bold}$ and second best {\ul underlined}. Arrows indicate the direction of improvement for each metric.}
    \begin{tabular}{@{}ccccc@{}}
    \toprule
    Exp. Nº & Method &  MAE ($\text{\AA}^2$)↓ & $S_{12}$ (\%)↓  & IoU (\%)↑ \\ \midrule
    1 & CartNet &   $\mathbf{2.88 \cdot 10^{-3}}$ & \textbf{0.75}  & \textbf{83.53}   \\
    2 & w/o hydrogens &  $3.28 \cdot 10^{-3}$ & 0.94 & 81.74  \\
    3 & w/o envelope &  $3.04 \cdot 10^{-3}$ & {\ul 0.77} & {\ul 83.21}           \\
    4 & w/o $\hat{\mathbf{v}}_{ij}$            & $6.23 \cdot 10^{-3}$ & 2.46 & 74.17           \\
    5 & w/o temperature     & $3.04 \cdot 10^{-3}$ & 0.85  & 82.22           \\
    6 & w/o SO(3) Aug  &  {\ul $3.02 \cdot 10^{-3}$} & 0.81 & 82.78           \\ 
    \bottomrule
    \end{tabular}
    
    \label{tab:ablations}
\end{table}

\subsection{Impact of the Hydrogens}

The ADP dataset used in this study does not contain experimental ADP data for the hydrogen atoms, which makes determining how to handle these atoms during the graph construction process challenging. 
Our proposed methodology takes advantage of the available 3D coordinates of the hydrogen atoms by including them as additional atoms in the graph but inferring ellipsoids solely for the non-hydrogen atoms.
However, we also tested an approach that ignored all hydrogen atoms during the creation of the graph and computed the ellipsoids using only the remaining atoms.

As seen in the experiment number 2 from the Table \ref{tab:ablations}, the results indicate that excluding the hydrogen atoms decreases the IoU from $83.53\%$ to $81.74\%$, resulting in a $1.32\%$ decrease in performance.
For the other metrics, an improvement of $14.89\%$ can be seen in MAE and $0.19\%$ in $S_{12}$.

The results suggest that hydrogen atoms, while not directly involved in inferring ellipsoids, contribute valuable contextual information to the graph. 
This additional data benefits the model, allowing it to more accurately infer the ADPs of non-hydrogen atoms by incorporating the effects of the covalent bonded hydrogens and include the relevant hydrogen bond interactions and other intermolecular forces.

\subsection{Neighbour Equalization}

The neighbour equalization technique is a novel method for equalizing the number of neighbours. 
It uses an envelope function to weight the contribuions of atoms with respect to their distance.
It addresses the challenge of distant atoms disproportionately influencing the aggregation process and helps detect the peaks of the different interatomic interactions.

Experiment number 3 in Table \ref{tab:ablations} shows the results when training CartNet without the envelope function.
Compared to the full CartNet (Experiment Nº 1), the ADPs prediction experiments a drop of $0.32\%$ for the IoU, $0.02\%$ for the $S_{12}$, and $5.26\%$ for the MAE.
This suggests that using neighbour equalization with the envelope function is highly effective in equalizing the neighbours.

\subsection{Cartesian Axis}

This ablation study investigates the effect of incorporating the Cartesian direction vector ($\hat{\mathbf{v}}_{ij}$) between atoms in the edge encoder on the CartNet model. 
The direction vector captures important geometric information about the relative orientation of atoms, which is expected to influence the accuracy of ADP predictions.

The results presented in experiment number 4 in Table \ref{tab:ablations} clearly show the significant impact of incorporating the Cartesian direction unit vector in the edge encoder. 
The $9.36\%$ decrease in IoU, the $1.71\%$ drop in the $S_{12}$, and the $53.77\%$ reduction in MAE highlight the ability of the model to capture the spatial relationships between atoms when using the direction vector.
This suggests that the direction unit vector is a crucial input feature for effectively encoding the geometric information necessary for accurate ADP prediction.

\subsection{Temperature}

This ablation study explores the impact of including temperature as an input feature on the performance of the model in predicting ADPs. 
Temperature plays a crucial role in atomic displacements, and incorporating it as an input feature can enhance the ability of the model to capture temperature-dependent behaviours in ADPs.

The experiment 5 in Table \ref{tab:ablations} shows the results when not including the temperature information to the input of the CartNet model.
The decrease $1.31\%$ in IoU, $0.08\%$ in the $S_{12}$, and $4.63\%$ in MAE demonstrate that the ability of the model to capture the spatial extent of atomic displacements is enhanced when the temperature is included.  
These results suggest that temperature is essential for improving the performance of models in predicting ADPs, particularly when considering the thermal motion of atoms.

\subsection{SO(3) Data Augmentation}
\label{sec:data-aug}

The augmentation was explicitly designed to enhance the model’s ability to learn to generalize unseen rotations, which is critical when dealing with 3D molecular and crystal structures where the orientation of the input can vary.

For these experiments, we trained our model on the ADP dataset with and without rotation SO(3) augmentation. 
The configuration without augmentation exposed the model to the data in its original orientation.
In contrast, the configuration utilizing rotation SO(3) augmentation involved randomly applying three-dimensional rotations to the direction vectors of the atoms during training, thereby encouraging the model to learn features equivariant to spatial orientation.

Experiment number 6, compared with experiment number 1 from Table \ref{tab:ablations}, demonstrates the positive impact of the rotation SO(3) augmentation on model performance. 
The decrease of $0.75\%$ in IoU,  $0.06\%$ in $S_{12}$, an $4.63\%$ in MAE demonstrates that the model is more adept at generalizing to unseen orientations of the input data, making it better suited for real-world applications where molecules and crystal structures can appear in various spatial configurations.

Since CartNet does not explicitly enforce SO(3) rotation equivariance but learns it through data augmentation, we wanted to evaluate the error this method introduces to the final predictions.
To quantify this, we defined two variables: $\mathbf{U}_{\text{orig}}$ represents the ADP predictions from the original, unrotated crystal structures, and $\mathbf{U}_{\text{rot}}$ represents the ADP predictions from the crystal structures after they have been rotated by a rotation matrix $\mathbf{R}$.
We then compared $\mathbf{U}_{\text{rot}}$ with the rotated versions of $\mathbf{U}_{\text{orig}}$, calculated as $\mathbf{R} \mathbf{U}_{\text{orig}} \mathbf{R}^\top$.
This methodology allows us to isolate rotation-induced errors, ensuring that any observed discrepancies are attributable solely to rotation effects rather than a combination of prediction and rotation errors.
To perform this comparison, we conducted a Monte Carlo experiment, applying 100 different random rotation matrices to the test set.

\begin{figure}[htb]
    \centering
    \includegraphics[width=0.5\linewidth]{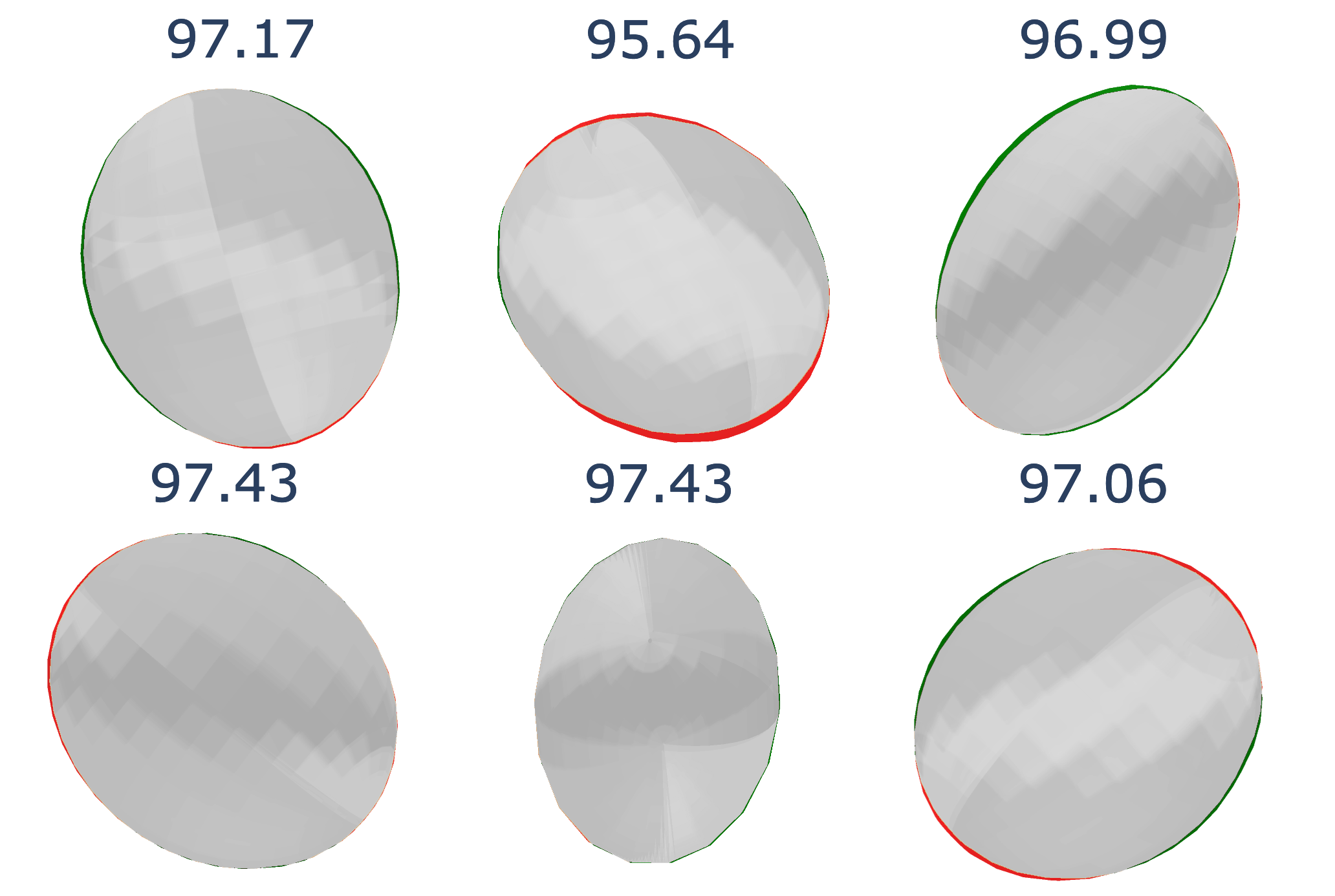}
    \caption{Visualization of rotational errors by comparing the rotated ADP predictions from the original crystal structures (green ellipsoids, $\mathbf{R} \mathbf{U}_{\text{orig}} \mathbf{R}^\top$) with the ADP predictions from the rotated crystal structures (red ellipsoids, $\mathbf{U}_{\text{rot}}$). The overlapping regions are shaded in grey, representing the intersection between the two predictions. The IoU values are displayed above each pair of ellipsoids.}
    \label{fig:montecarlo}
\end{figure}

The results of this experiment yielded a Mean Absolute Error (MAE) of $1.01 \times 10^{-3}\,\text{\AA}^2 \pm 1.68 \times 10^{-3}\,\text{\AA}^2$, an $S_{12}$ score of $0.65\% \pm 0.17\%$, and an Intersection over Union (IoU) of $94.96\% \pm 2.46\%$.
Figure~\ref{fig:montecarlo} illustrates the results of this experiment.

In all cases, the ADP predictions for the rotated crystal structures ($\mathbf{U}_{\text{rot}}$) closely matched the rotated predictions of the original structures ($\mathbf{R} \mathbf{U}_{\text{orig}} \mathbf{R}^\top$).
These results confirm that CartNet effectively generalizes to unseen rotations despite not explicitly enforcing rotation equivariance.

\section{Conclusions}

In this work, we introduced CartNet, a novel GNN architecture designed to predict properties of crystalline molecular-based structures. 
In the specific test case examined, our model significantly reduces computational costs while demonstrating improved performance relative to the low-level GGA functional DFT calculations and current state-of-the-art learning-based architectures. 
Nonetheless, more advanced DFT formulations (e.g., hybrid functionals with many-body dispersion), which are considerably more computationally expensive, may offer higher accuracy and warrant further investigation to fully assess the broader performance benefits of our approach.
The development of CartNet was driven by the need to address the challenges posed by ADPs, which are crucial for understanding thermal vibrations in crystallography.

CartNet utilizes a novel Cartesian encoding approach that avoids reliance on the unit cell, thereby overcoming limitations faced by previous models.
The incorporation of neighbour equalization helps the model to differentiate between various types of bonds and interaction forces between atoms.
The Cholesky-based output layer ensures the model generates valid ADP predictions that align with physical requirements. 
Additionally, by introducing a rotational generalitzation through data augmentation, CartNet effectively learns the directional nature of atomic vibrations without relying on specific equivariant layers.
The evaluation of CartNet demonstrated its robustness and accuracy. 
It outperformed previously reported methods in other benchmarks, Jarvis and The Materials Project, that focused on bulk materials, instead of molecular systems, and contained structure and properties that have been computed with DFT calculations, instead of experimental structures and ADPs.

In addition to the model, we curated and presented a comprehensive ADP dataset containing over $200\text{k}$ crystal structures of molecular systems from the Cambridge Structural Database (CSD). 
This dataset spans a wide range of temperatures and atomic environments, providing a valuable resource for further research on predicting anisotropic displacements and thermal behaviours in crystalline structures.

This work provides a more efficient and accurate method for predicting properties in crystal structures, opening new possibilities for studying different material properties and designing new materials.
Therefore, when CartNet is specifically used to predict ADPs, it can be used to evaluate the experimental results of new systems in cases where their experimental determination using diffraction techniques presents difficulties.

Future work could focus on predicting cell expansion to estimate ellipsoids at other temperatures based on a fixed geometry at a specific temperature.
Regarding the specific case of the ADP, future work could explore the creation of equivariant methods to generate valid ADP matrices.
Moreover, due to the large number of molecular crystal structures in the ADP dataset, future work can study using this dataset as pre-training for other crystal structure tasks.
Finally, the efficiency and accuracy of CartNet also highlight its potential for future crystal structure prediction challenges, such as the 7th Blind Test~\cite{seventhccdcchallenge,seventhccdcchallenge_2} organized by the Cambridge Crystallographic Data Centre (CCDC)~\cite{CCDC}, where handling highly flexible or disordered systems remains a critical obstacle.

\section*{Author contributions}

\textbf{Àlex Solé}: Data curation, Conceptualization, Methodology, Software, Investigation, Validation, Writing - review \& editing. 
\textbf{Albert Mosella-Montoro}: Conceptualization, Supervision, Resources, Writing - review \& editing.
\textbf{Joan Cardona}: Conceptualization, Writing - review \& editing.
\textbf{Silvia Gómez-Coca}: Conceptualization, Supervision, Resources, Writing - review \& editing.
\textbf{Daniel Aravena}: Conceptualization, Supervision, Resources, Validation,  Writing - review \& editing.
\textbf{Eliseo Ruiz}: Conceptualization, Supervision, Resources, Writing - review \& editing, Funding acquisition.
\textbf{Javier Ruiz-Hidalgo}: Conceptualization, Supervision, Resources, Writing - review \& editing, Funding acquisition.

\section*{Conflicts of interest}
There are no conflicts to declare.

\section*{Data availability}

The code to generate the ADP Dataset and recreate the results of the paper can be found at:
\href{https://github.com/imatge-upc/CartNet}{https://github.com/imatge-upc/CartNet}.
Project website with online demo available at: \href{https://www.ee.ub.edu/cartnet}{https://www.ee.ub.edu/cartnet}.

\section*{Acknowledgements}

Financial support from Ministerio de Ciencia e Innovación (project  PID2020-117142GB-I00, PID2021-122464NB-I00, TED2021-129593B-I00, CNS2023-144561 and Maria de Maeztu CEX2021-001202-M). We also acknowledge the Generalitat de Catalunya for the 2021-SGR-00286 grant. E.R. also acknowledges the Generalitat de Catalunya for an ICREA Academia grant. We thank BSC for the computational resources.

\bibliographystyle{rsc}  
\bibliography{refs.bib}

\appendix
\appendixpage

\renewcommand{\thesection}{S\arabic{section}}
\setcounter{section}{0}

\renewcommand{\thefigure}{S\arabic{figure}}
\setcounter{figure}{0}
\renewcommand{\thetable}{S\arabic{table}}
\setcounter{table}{0}

\section{Envelope}

\begin{figure}[htb]
    \centering
    \includegraphics[width=0.5\linewidth]{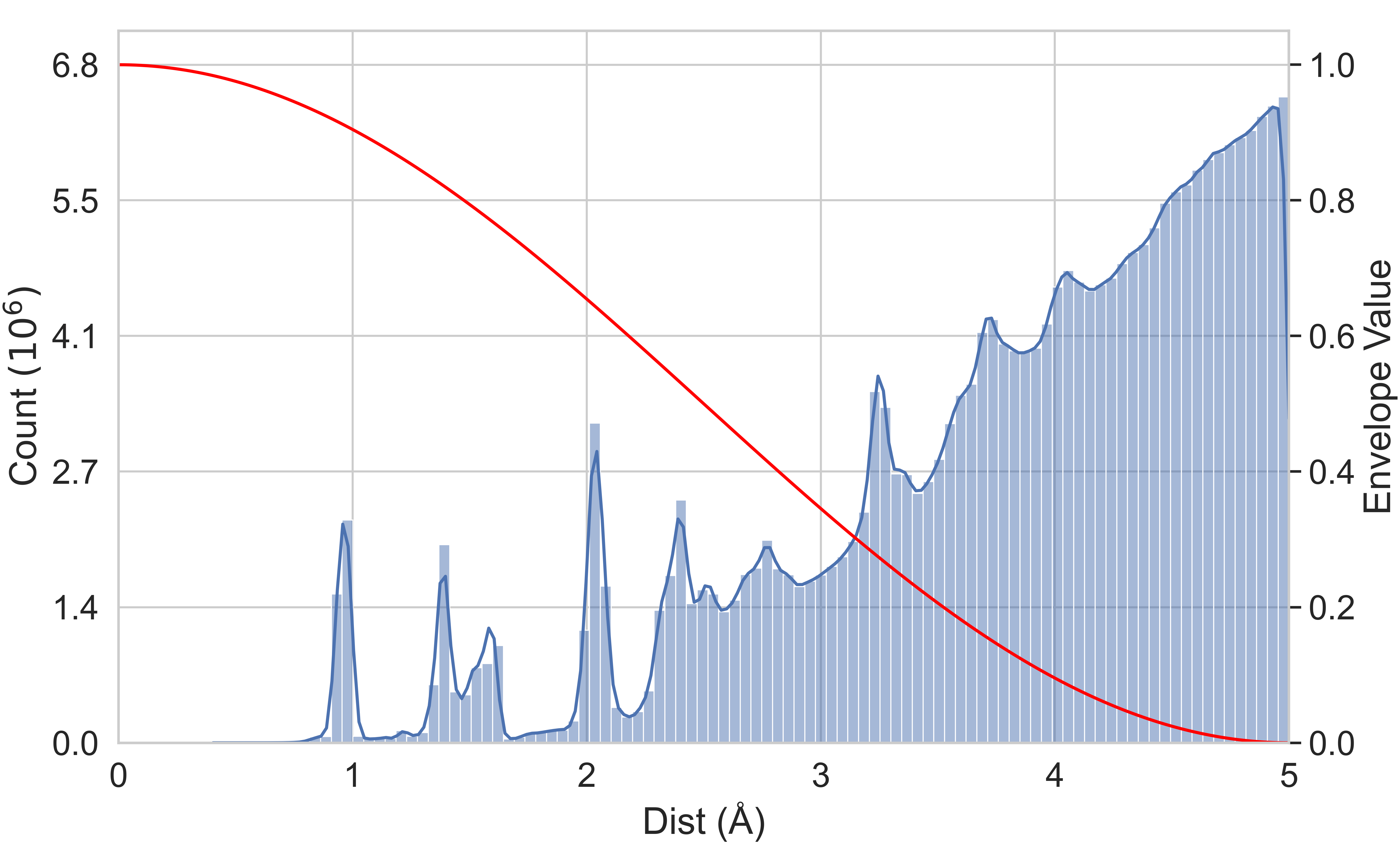}
    \caption{Histogram illustrating the distribution of edges in the test split of the ADP dataset. In blue, the distribution of the number of edges (displayed on the left y-axis) as a function of the distance between atoms in \AA{}. The red curve represents the Envelope over distance, plotted on the right y-axis.}
    \label{fig:envelope}
\end{figure}

Figure \ref{fig:envelope} shows the Envelope function compared with the histogram of the number of edges in the test set from the ADP dataset.
The histogram is consistent with the distribution of interatomic distances described by previous works~\cite{santiago}. 
Showing covalent peaks with multiple hits at short distances, and also a large number of long-distance peaks related to contacts.
Further confirming that weaker intermolecular interactions play a significant role in our dataset.

\section{Rotation SO(3) ADP}

Given $\mathbf{U_i}$ as the covariance matrix of atom $i$, $\mathbf{U_i}$  can be decomposed using the expected value of a set of N random samples, as defined in Equation (\ref{eq:U_decom}).

\begin{equation}
    \mathbf{U_i} = \mathbb{E}\left(\mathbf{X X}^{ T}\right)-\mathbb{E}\left(\mathbf{X}\right) \mathbb{E}\left(\mathbf{X}^{T}\right) \in \mathbb{R}^{3\times3}
    \label{eq:U_decom}
\end{equation}

Where $\mathbf{X}$ is a matrix with the coordinates of the N random samples, such that $\mathbf{X} \in \mathbb{R}^{N \times 3}$. 
The rotated version of $\mathbf{X}$ is given by $\mathbf{X}^{\prime}=\mathbf{R X}$.
The rotated covariance matrix, therefore, can be derived as Equation (\ref{eq:rot_U}).

\begin{equation}
\begin{aligned}
\mathbf{U^{aug}_i} & =\mathbb{E}\left(\mathbf{X}^{\prime} \mathbf{X}^{\prime T}\right)-\mathbb{E}\left(\mathbf{X}^{\prime}\right) \mathbb{E}\left(\mathbf{X}^{\prime T}\right) \\
& =\mathbb{E}\left(\mathbf{R X X}^T \mathbf{R}^T\right)-\mathbb{E}(\mathbf{R X}) \mathbb{E}\left(\mathbf{X}^T \mathbf{R}^T\right) \\
& =\mathbf{R} \mathbb{E}\left(\mathbf{X X}^T\right) \mathbf{R}^T-\mathbf{R} \mathbb{E}(\mathbf{X}) \mathbb{E}\left(\mathbf{X}^T\right) \mathbf{R}^T \\
& =\mathbf{R}\left(\mathbb{E}\left(\mathbf{X X}^T\right)-\mathbb{E}(\mathbf{X}) \mathbb{E}\left(\mathbf{X}^T\right)\right) \mathbf{R}^T \\
& =\mathbf{R} \mathbf{U_i}  \mathbf{R^T} \in \mathbb{R}^{3\times3}
\end{aligned}
\label{eq:rot_U}
\end{equation}

This derivation demonstrates how the covariance matrix $\mathbf{U_i}$ is transformed under rotation, ensuring that the rotated ellipsoid maintains its correct orientation relative to the input data.

\section{Intersection over Union (IoU) computation}
\label{sec:iou_compt}

The Intersection over Union (IoU) of two ADPs is not straightforward to compute analytically. 
To address this, the 3D space defined by $[-1,1]^3$ is discretized into a voxel grid with dimensions $64 \times 64 \times 64$.
To ensure that both the true and predicted ellipsoids are confined within the $[-1,1]^3$ space, the ellipsoids are normalized by the maximum norm between the true and predicted ellipsoid matrices $\mathbf{U}$.
The Mahalanobis distance is computed for each voxel relative to the predicted and ground truth ellipsoids.
The Mahalanobis distance measures the distance between a point and a distribution and is given by the Equation \ref{eq:iou}

\begin{equation}
D_M(\mathbf{x}) = \sqrt{\mathbf{x}^T \mathbf{U_{norm}}^{-1} \mathbf{x}}
\label{eq:iou}
\end{equation}

Where $\mathbf{U_{norm}}$ is defined as $\frac{\mathbf{U}}{\max(\|\mathbf{U}_{\text{pred}}\|, \|\mathbf{U}_{\text{true}}\|)}$, and $\mathbf{x}$ represents the coordinates of the voxel center.

\begin{figure}[ht]
\centering
\includegraphics[width=0.7\textwidth]{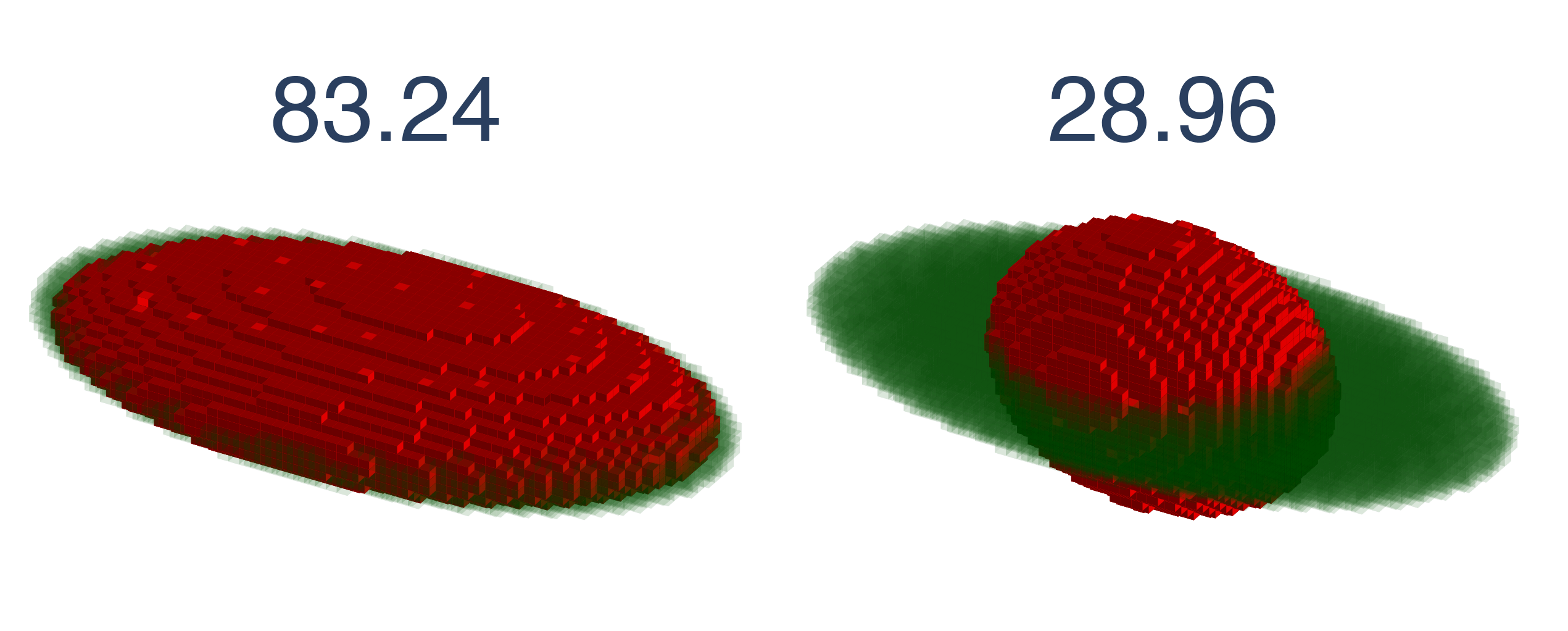}
\caption{Visual examples of the voxelization method and the respective IoU. The green regions represent the ground truth values and the red ones the predicted. The green regions have been plotted semitransparent to facilitate the visualization of the intersection between the two regions.}
\label{fig:minecraftlanobis}
\end{figure}

The resulting distance map is then binarized using a threshold of 1, such that voxels with $D_M(\mathbf{x}) \leq 1$ are considered inside the ellipsoid, while the rest are considered outside.
The IoU is subsequently computed using the binarized distance maps for predicted and ground truth ellipsoids.
Two examples of the binarized voxel grid and their respective IoU can be seen in Figure \ref{fig:minecraftlanobis}.

\section{Training Details}

In this section, we outline the specific training configurations employed for each dataset.
For all the trainings, Adam~\cite{kingma2014adam} optimizer with OneCycle~\cite{smith2019super} scheduler with pct start of 0.01 policy has been used.
For the Jarvis and the Material Project dataset, we modified the head of the CartNet to be able to predict a scalar value ($s$).
The prediction head, $MLP_{head} $, replaces the Cholesky head for these two datasets. It consists of two linear layers: the first reduces the dimensionality from $ dim $ to $ dim/2 $, followed by a SiLU activation function, and the second further reduces the dimensionality from $ dim/2 $ to $ 1 $. 
The final output is obtained by applying $ MLP_{head} $ to each latent vector $ h^L_i $ and then averaging the resulting scalar values across all nodes within the unit cell, denoted by $ AVG $.
Equation (\ref{eq:scalar_head}) defines mathematically the architecture used.

\begin{equation}
    s = AVG(MLP_{head}(h^L_{i}))
    \label{eq:scalar_head}
\end{equation}

Where $ h^L_i $ denote the latent vector at the final CartNet layer for atom $ i $.

\subsection{ADP Dataset}
\label{sec:train_config_ellipsoid}
The configuration used for training the different state-of-the art models in the ADP Dataset can be found in Table \ref{tab:config_ellip}.

\begin{table}[h]
\centering
\caption{Training configuration used for the ADP Dataset}
\resizebox{\textwidth}{!}{%
\begin{tabular}{@{}cccccccccc@{}}
\toprule
Method     & \#Layers & Embedding Dim & Batch Size & Batch Accumulation & Lr                                                & Epochs & Loss & SO(3) Augmentation & Neigh. Strategy \\ \midrule
iComformer & 4        & 256           & 4          & 16                 & $1 \cdot 10^{-3}$ & 50     & L1   & False      & KNN-25          \\
eComformer & 4        & 256           & 4          & 16                 & $1 \cdot 10^{-3}$ & 50     & L1   & False      & KNN-25          \\
Ours & 4 & 256 & 4 & 16 & $1 \cdot 10^{-3}$ & 50 & L1 & True & \begin{tabular}[c]{@{}c@{}}Radius Graph\\ (5~\AA{})\end{tabular} \\ \bottomrule
\end{tabular}
}

\label{tab:config_ellip}
\end{table}

\subsection{Jarvis Dataset}
\label{sec:train_config_jarvis}
The specific configurations used for training the different properties the Jarvis dataset provides are detailed in Table \ref{tab:config_jarvis}.

\begin{table}[H]
\centering
\caption{Training configuration used for the Jarvis dataset.}
\resizebox{\textwidth}{!}{%
\begin{tabular}{@{}cccccccccc@{}}
\toprule
Property & \#Layers & Embedding Dim & Batch Size & Batch Accumulation & Lr & Epochs & Loss & SO(3) Augmentation & Neigh. Strategy \\ \midrule
Formation Energy & 4 & 256 & 64 & 1 &  $1 \cdot 10^{-3}$ & 500 & L1 & False & \begin{tabular}[c]{@{}c@{}}Radius Graph\\ (5~\AA{})\end{tabular} \\
Band Gap (OPT)    & 4 & 256 & 64 & 1 &  $1 \cdot 10^{-3}$  & 500 & L1 & False & \begin{tabular}[c]{@{}c@{}}Radius Graph\\ (5~\AA{})\end{tabular} \\
Total Energy     & 4 & 256 & 64 & 1 &  $1 \cdot 10^{-3}$ & 500 & L1 & False & \begin{tabular}[c]{@{}c@{}}Radius Graph\\ (5~\AA{})\end{tabular} \\
Band Gap (MBJ)    & 4 & 256 & 64 & 1 &  $1 \cdot 10^{-3}$  & 500 & L1 & False & \begin{tabular}[c]{@{}c@{}}Radius Graph\\ (5~\AA{})\end{tabular} \\
Ehull            & 4 & 256 & 64 & 1 &  $1 \cdot 10^{-3}$ & 500 & L1 & False & \begin{tabular}[c]{@{}c@{}}Radius Graph\\ (5~\AA{})\end{tabular}  \\ \bottomrule
\end{tabular}%
}

\label{tab:config_jarvis}
\end{table}

\subsection{The Materials Project Dataset}
\label{sec:train_config_megnet}

The specific configurations used for training the different properties the Materials Project dataset provides are detailed in Table \ref{tab:config_megnet}.

\begin{table}[H]
\centering
\caption{Training configuration used for the Materials Project dataset.}
\resizebox{\textwidth}{!}{%
\begin{tabular}{@{}cccccccccc@{}}
\toprule
Property         & \#Layers & Embedding Dim & Batch Size & Batch Accumulation & Lr                             & Epochs & Loss & SO(3) Augmentation & Neigh. Strategy \\ \midrule
Formation Energy & 4        & 256           & 64         & 1                  &  $1 \cdot 10^{-3}$ & 500    & L1   & False  & \begin{tabular}[c]{@{}c@{}}Radius Graph\\ (5~\AA{})\end{tabular}             \\
Band Gap         & 4        & 256           & 64         & 1                  &  $1 \cdot 10^{-3}$ & 500    & L1   & False  & \begin{tabular}[c]{@{}c@{}}Radius Graph\\ (5~\AA{})\end{tabular}             \\
Form. Energy         & 4        & 256           & 64         & 1                  &  $1 \cdot 10^{-3}$ & 500    & L1   & False  & \begin{tabular}[c]{@{}c@{}}Radius Graph\\ (5~\AA{})\end{tabular}             \\
Band Gap        & 4        & 256           & 64         & 1                  & $1 \cdot 10^{-3}$ & 500    & L1   & False  & \begin{tabular}[c]{@{}c@{}}Radius Graph\\ (5~\AA{})\end{tabular}             \\ \bottomrule
\end{tabular}%
}

\label{tab:config_megnet}
\end{table}

\section{Results Analysis}

In addition we have evaluated the ADPs and the performance of our model for several chemical interactions or atom types. Table~\ref{tab:interactions} summarizes these results. In addition to the MAE, $S_{12}$, and the IoU, we include the volume of the experimental ellipsoid that provides a reference for the typical spatial extent associated with each atom type. For comparison purposes, the metrics for all the atoms and only carbon atoms have been included as "Any" and C, respectively. For this analysis, two typical intermolecular interactions have been considered, hydrogen bonds and $\pi$-$\pi$-interactions between benzene rings. For H-bonds the definition employed by CSD has been used and for $\pi$-$\pi$-interactions only benzene rings with centroids at a distance between 3 and 4 \text{\AA} have been considered.

\begin{table}[h]
  \centering
  \caption{Analysis of the results for different atom types of the ADP test dataset. Arrows indicate the direction of improvement for each metric.}
  \begin{tabular}{ccccc}
    \hline
    Atom type     & Volume ($\text{\AA}^3$) &  MAE ($\text{\AA}^2$)↓ & $S_{12}$ (\%)↓  & IoU (\%)↑   \\ \hline
    Any & $3.08 \cdot 10^{-2}$ & $2.88 \cdot 10^{-3} $ &  $0.75$ & $83.53 $ \\
    C & $3.12 \cdot 10^{-2} $ & $2.90 \cdot 10^{-3} $ & $0.74$ & $83.70$ \\
    $R=N,O,S$ in H-bond ($R-H \cdots R^\prime$)  & $2.52 \cdot 10^{-2}$ & $2.65 \cdot 10^{-3} $ & $0.89$ & $82.14 $ \\
    $R^\prime=N,O,F,Cl,Br,I$ in H-bond ($R-H \cdots R^\prime$)  & $2.76 \cdot 10^{-2}$ & $2.75 \cdot 10^{-3} $ & 0.83 & $82.78 $ \\
    C in $\pi - \pi$ interaction & $2.76 \cdot 10^{-2}$ &  $  2.43 \cdot 10^{-3} $ & $0.66$ & $85.54$ \\
    central C in tert-Butyl  & $2.23 \cdot 10^{-2}$ & $2.33 \cdot 10^{-3} $ & $0.78$ & $82.80$ \\
    methyl C in tert-Butyl & $3.84 \cdot 10^{-2}$ & $3.81 \cdot 10^{-3} $ &  $0.73$ & $83.60$ \\
    \hline
  \end{tabular}
  \label{tab:interactions}
\end{table}

For atoms participating in hydrogen bonds, moderate MAEs ($\sim2.65 \times 10^{-3}\,\text{\AA}^2$ to $2.75 \times 10^{-3}\,\text{\AA}^2$) are observed, alongside slightly elevated $S_{12}$ values (0.89 and 0.83). In these atoms the experimental volumes ($\sim2.52 \times 10^{-2}\,\text{\AA}^3$ to $2.76 \times 10^{-2}\,\text{\AA}^3$) are smaller than all the atom types ($\sim3.08 \times 10^{-2}\,\text{\AA}^3$) indicating some spatial constraints as expected, even at moderate level. Beside moderates, the IoU values around 82\% confirm that the predicted ellipsoids still overlap well with the reference data.

In the case of $\pi$-$\pi$-interactions between benzene rings, it stand out with an MAE of $2.43 \times 10^{-3}\,\text{\AA}^2$, which is lower than that of the general carbon category ($2.90 \times 10^{-3}\,\text{\AA}^2$). This reduced MAE aligns with their smaller average experimental volume of $2.76 \times 10^{-2}\,\text{\AA}^3$, compared with $3.12 \times 10^{-2}\,\text{\AA}^3$ for generic carbons. In other words, the ADPs in benzene rings are smaller than in other C atom types, probably due to the ring rigidity and reduced conformational flexibility. Moreover, the IoU of 85.54\% for these $\pi$-$\pi$-interacting carbons exceeds both the baseline for all atoms (83.53\%) and the baseline for carbons atoms (83.70\%), indicating that the model recovers the shape of their displacement ellipsoids particularly well. 

Additionally, the performance of the model can be affected by substituents that are more prone to disorder due to different factors, for example, more rotational freedom. One case scenario can be the tert-butyl group, that is the one analyzed here. It has two types of C atoms, the central one that is expected to have less freedom and the methyl ones that are expected to have more flexibility and consequently larger ADPs. As expected, the central carbon in the tert-butyl group,  exhibits one of the smallest experimental volumes among all carbons analyzed ($2.23 \times 10^{-2}\,\text{\AA}^3$). Correspondingly, it shows a lower MAE ($2.33 \times 10^{-3}\,\text{\AA}^2$) compared to both the “Any” and generic carbon categories. This observation supports the notion that a smaller, more spatially confined environment correlates with more accurate ADP predictions. Although the IoU for this central carbon (82.80\%) is similar to that of hydrogen-bonded atoms, it remains close to the baseline levels, indicating that the model is capturing the overall shape of the anisotropic displacement despite slight nuances in the ellipsoid’s geometry. Not surprisingly, the methyl carbons in tert-butyl groups occupy a larger experimental volume ($3.84 \times 10^{-2}\,\text{\AA}^3$) and show a higher MAE ($3.81 \times 10^{-3}\,\text{\AA}^2$) than any of the listed categories. However, their IoU (83.60\%) remains comparable to generic carbons, indicating that the model consistently identifies the shape of the ellipsoid, and is able to properly reproduce the increased thermal motion and conformational freedom in these methyl carbons.

Overall, these results highlight the robust predictive capability of our model across a variety of atom types and interactions. In addition, the correlation between the volume and the accuracy metrics (particularly MAE) emphasizes the influence of local structural constraints, such as planarity, rigidity, or conformational freedom, on anisotropic displacement parameters.

In order to gain more insight into how the model performs in the same compound but with different crystal structures and measured at different temperatures the results for the structures of the known ROY polymorph that were in the test dataset have been analyzed. Table~\ref{tab:polymorph} summarizes the performance of CartNet in predicting their ADPs. The metrics examined (MAE, $S_{12}$, and IoU) provide complementary insights into how faithfully the model captures thermal ellipsoid shapes and orientations. 

Overall, the model demonstrates robust predictive capability for the majority of the structures, often achieving high IoU values and low \(S_{12}\) and MAE metrics. Notably, several Y polymorphs (e.g., QAXMEH22, QAXMEH23, QAXMEH58) exhibit both high IoU scores (above 88\%) and minimal shape discrepancy, suggesting that CartNet can effectively capture thermal motion when the underlying structural motifs remain consistent. However, a performance reduction is observed for certain polymorphs measured at low temperatures or within the \(P\bar{1}\) space group. This trend is exemplified by QAXMEH53 (Y04, \(P\bar{1}\) at 100\,K) and QAXMEH56 (R, \(P\bar{1}\) at 150\,K), which show relatively higher \(S_{12}\) values alongside lower IoU scores. Similarly, although QAXMEH19 (Y, \(P2_1/n\) at 30\,K) maintains a comparatively low MAE, its IoU is notably lower than other Y polymorphs at higher temperatures, indicating that temperature-induced lattice distortions or pronounced packing differences can pose additional challenges for the model. 

\begin{table}[h]
\centering
\caption{Table of results between the different ROY structures present in the ADP test dataset. 
Arrows indicate the direction of improvement for each metric.}
\begin{tabular}{ccccccc}
\toprule
CSD Refcode & Polymorph & Space Group & Temperature (K) & MAE ($\text{\AA}^2$)↓ & $S_{12}$ (\%)↓  & IoU (\%)↑   \\
\midrule
QAXMEH32 & ON   & P2$_1$/c & 100 & $1.21  \cdot 10^{-3}$ & 0.56 & 80.95 \\
QAXMEH54 & ON   & P2$_1$/c & 150 & $2.09  \cdot 10^{-3}$ & 0.44 & 84.44 \\
QAXMEH55 & ORP  & Pbca     & 150 & $1.74  \cdot 10^{-3}$ & 0.29 & 88.50 \\
QAXMEH56 & R    & P$\bar{1}$ & 150 & $3.98  \cdot 10^{-3}$ & 1.71 & 68.46 \\
QAXMEH19 & Y    & P2$_1$/n & 30  & $1.26  \cdot 10^{-3}$ & 1.22 & 72.89 \\
QAXMEH22 & Y    & P2$_1$/n & 293 & $2.56  \cdot 10^{-3}$ & 0.23 & 90.09 \\
QAXMEH23 & Y    & P2$_1$/n & 293 & $2.50  \cdot 10^{-3}$ & 0.27 & 88.56 \\
QAXMEH58 & Y    & P2$_1$/n & 150 & $1.31  \cdot 10^{-3}$ & 0.20 & 91.95 \\
QAXMEH53 & Y04  & P$\bar{1}$ & 100 & $2.75  \cdot 10^{-3}$ & 1.42 & 68.74 \\
QAXMEH12 & YT04 & P2$_1$/n & 296 & $2.14  \cdot 10^{-3}$ & 0.16 & 90.93 \\
\bottomrule
\end{tabular}
\label{tab:polymorph}
\end{table}

Low temperatures were anticipated to yield less accurate ADPs, as discussed in Section~5.2.3 of the main manuscript. Nonetheless, we also examined whether there is a correlation between the IoU metric and the space groups listed in Table~\ref{tab:polymorph}. To this end, Table~\ref{tab:spacegroup_summary} presents the number of structures in the ADP test dataset alongside the mean IoU value for each space group. No correlation was observed between the space groups and the IoU metric.

\begin{table}[h]
\centering
\caption{Summary of space groups, number of structures, and mean IoU percentages of the ADP test dataset. Arrows indicate the direction of improvement for each metric.}
\begin{tabular}{ccc}
\toprule
Space Group & Number of Structures & IoU (\%)↑ \\
\midrule
P21/c & 5327 & 84.29 \\
P-1   & 4247 & 84.15 \\
P21/n & 4125 & 83.89 \\
Pbca  & 1079 & 83.97 \\
\bottomrule
\end{tabular}

\label{tab:spacegroup_summary}
\end{table}

\section{DFT calculations}
\label{sec:phonopy-config}
Density functional theory (DFT) calculations were carried out employing the Vienna ab initio Simulation package (VASP v6.4.3~\cite{vasp1, vasp2, vasp3}. ADPs were calculated based on numerical displacements around the optimized geometry, generated using the Phonopy program~\cite{phonopy-phono3py-JPSJ, phonopy-phono3py-JPCM} (v2.19.1). All DFT calculations were based on the PBE~\cite{PBE} functional, the projector-augmented wave method ($E_{cut}$ = $500$ eV) and included D3-BJ~\cite{d3} dispersion corrections. Electron wave functions were converged to a threshold of $10^{-8}$ eV. Calculations were repeated for three different optimized geometries: (i) a fully relaxed structure, including atomic positions and unit cell parameters, (ii) a geometry where the unit cell was fixed to the crystallographic parameters, allowing the atomic positions to relax and (iii) a structural relaxation constrained to a fixed volume ($241.23687\,\text{\AA}^2$), calculated by the Vinet equation of state for a series of compressed and expanded unit cells using the quasi-harmonic approximation. The thermal expansion was calculated for $298\text{K}$ to match the experimental data from the CSD database for the ETIDEQ refcode. Structural relaxation runs considered a k-point grid of $5 {\times} 3 {\times} 2$ while the displaced geometries were calculated at the $\Gamma$ point using a $5 {\times} 3 {\times} 2$ supercell. ADPs were obtained using a q-point grid of $64 {\times} 64 {\times} 64$. Regarding the computational requirements, calculations for each displaced geometry took ca. $20{,}000$ s ($5.8$ hrs) to complete using $56$ cores in a parallel run. The fully relaxed calculation and the one derived from the Vinet equation required $144$ displacements to complete while the simulations based on the experimental unit cell demanded $72$ displacements, as it retained inversion symmetry. Six additional calculations, each including $144$ displacements, were done for the compressed and expanded units cells.

\section{Temperature Ablation}

The temperature ablations from the paper mention a series of crystal structures of \textit{guanidinium pyridiniumnaphthalene-1,5-disulfonate}. They are used to assess the ability of CartNet to predict ADPs at other temperatures using as input the crystal structure geometry obtained at a different temperature.
The refcodes used for the study with the respective temperature can be found in Table \ref{tab:temp_ref}.

\begin{table}[h]
\centering
\caption{The refcodes and the temperatures used for the temperature ablation study.}
\begin{tabular}{@{}cc@{}}
\toprule
Refcode & Temperature (K)               \\ \midrule
DOWVOC28 & 153 \\
DOWVOC31 & 163 \\
DOWVOC33 & 178 \\
DOWVOC34 & 183 \\
DOWVOC36 & 193 \\
DOWVOC38 & 203 \\
DOWVOC40 & 213 \\
DOWVOC42 & 223 \\
DOWVOC44 & 233 \\
DOWVOC46 & 243 \\
DOWVOC48 & 253 \\
DOWVOC02 & 263 \\
DOWVOC04 & 273 \\
DOWVOC29 & 283 \\ \bottomrule
\end{tabular}%

\label{tab:temp_ref}
\end{table}

\begin{figure}[htb]
  \centering
  \includegraphics[width=0.5\linewidth]{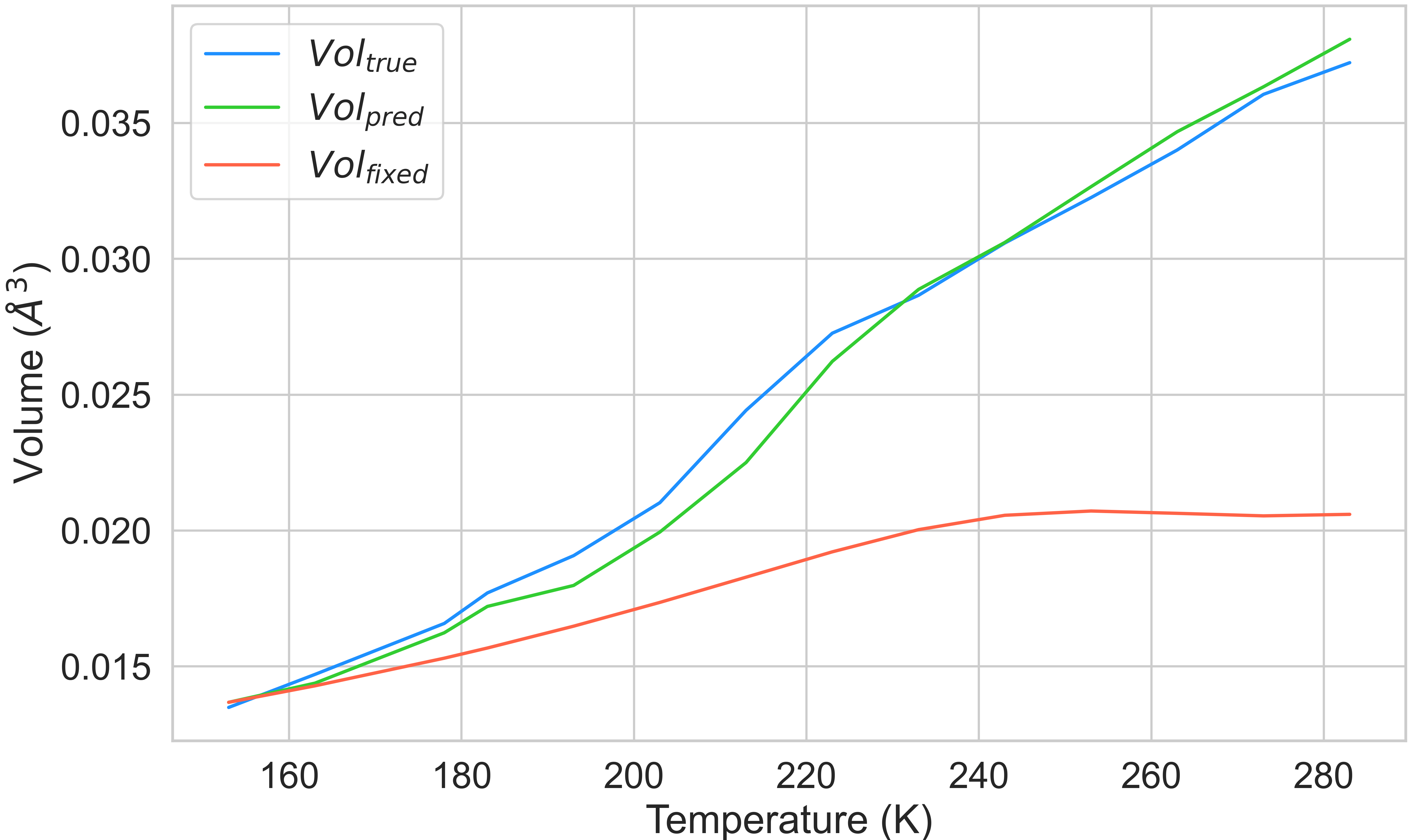}
  \captionof{figure}{Comparison of mean volume of the ellipsoids as a function of temperature for the \textit{guanidinium pyridinium naphthalene-1,5-disulfonate} (CSD refcode: DOWVOC) crystal structure. Blue line represent the experimental volumes, green line represent the predicted volume of the ellipsoid using the geometry and temperature from experimental data, and orange line represent the predicted volume of the ellipsoid when using the fixed geometry from the crystal structure at 153K and modifying the input temperature to CartNet.}
  \label{fig:anchor_min}
\end{figure}

\begin{figure}[htb]
  \centering
  \includegraphics[width=0.5\linewidth]{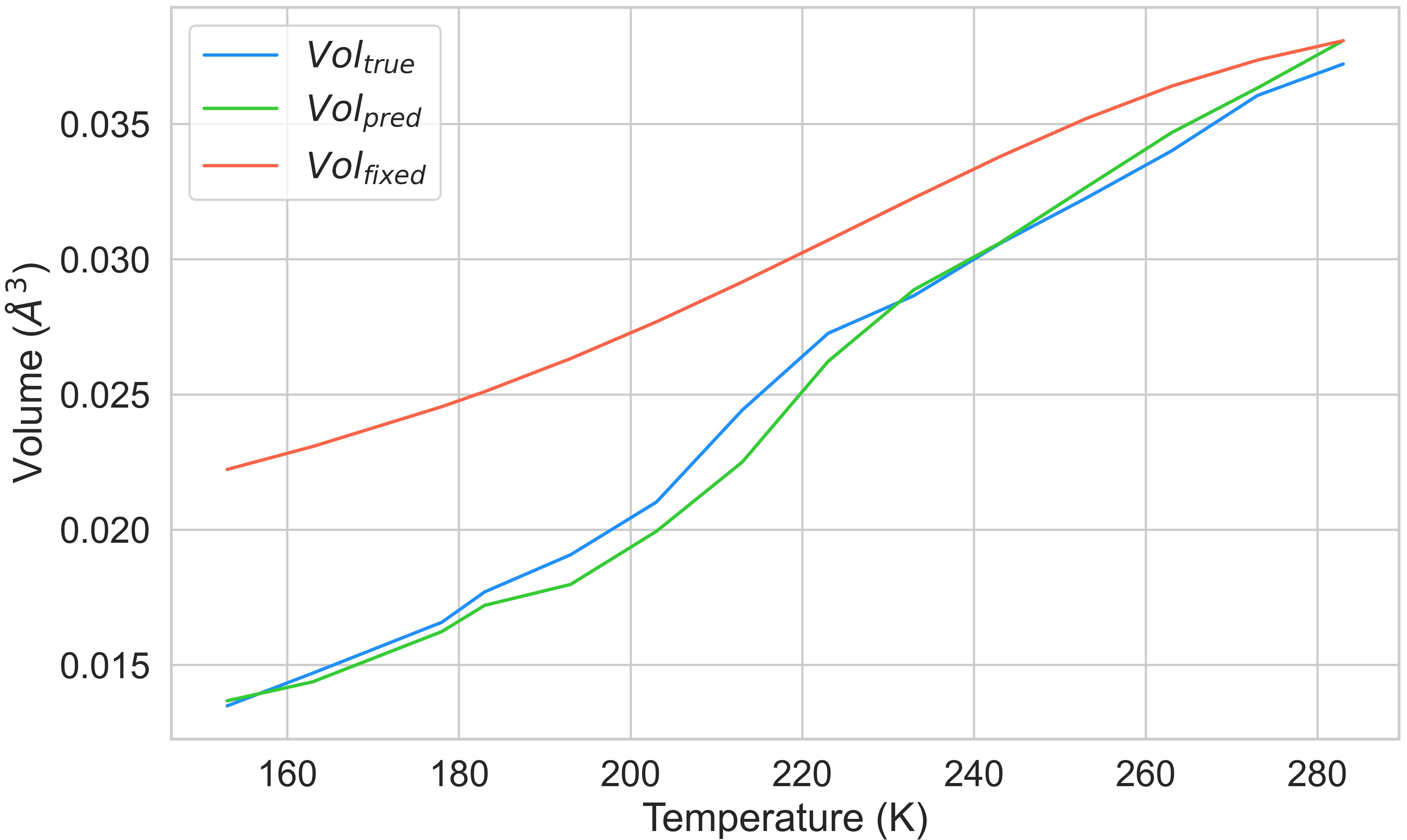}
  \captionof{figure}{Comparison of mean volume of the ellipsoids as a function of temperature for the \textit{guanidinium pyridinium naphthalene-1,5-disulfonate} (CSD refcode: DOWVOC) crystal structure. Blue line represent the experimental volumes, green line represent the predicted volume of the ellipsoid using the geometry and temperature from experimental data, and orange line represent the predicted volume of the ellipsoid when using the fixed geometry from the crystal structure at 283K and modifying the input temperature to CartNet.}
  \label{fig:anchor_max}
\end{figure}

We repeated the fixed-geometry experiment at 153~K and 283~K , adjusting only the input temperature to CartNet. Figure~\ref{fig:anchor_min} shows results obtained with the 153~K geometry, and Figure~\ref{fig:anchor_max} presents those from the 283~K geometry.
In both cases, deviations from experimental values grow as the input temperature moves further from the reference structure. This finding indicates that significant structural rearrangements (e.g., intermolecular interactions, phase transitions) cannot be fully captured when the geometry is constant. Although CartNet accounts for some thermal expansion through temperature input, it does not accommodate large-scale reorganizations without an updated geometry.
Overall, CartNet reasonably reproduces thermal ellipsoids but shows limitations when temperature changes are substantial enough to trigger phase transitions or significant structural differences.

\newpage

\section{Visual Results}
\label{sec:vis_res}

This section presents some more comparison results between the ADPs prediction of eConformer, iConformer and our proposed CartNet method. 

\subsection{DOLBIR21}

\begin{figure}[H]
    \centering
    \includegraphics[width=0.9\linewidth]{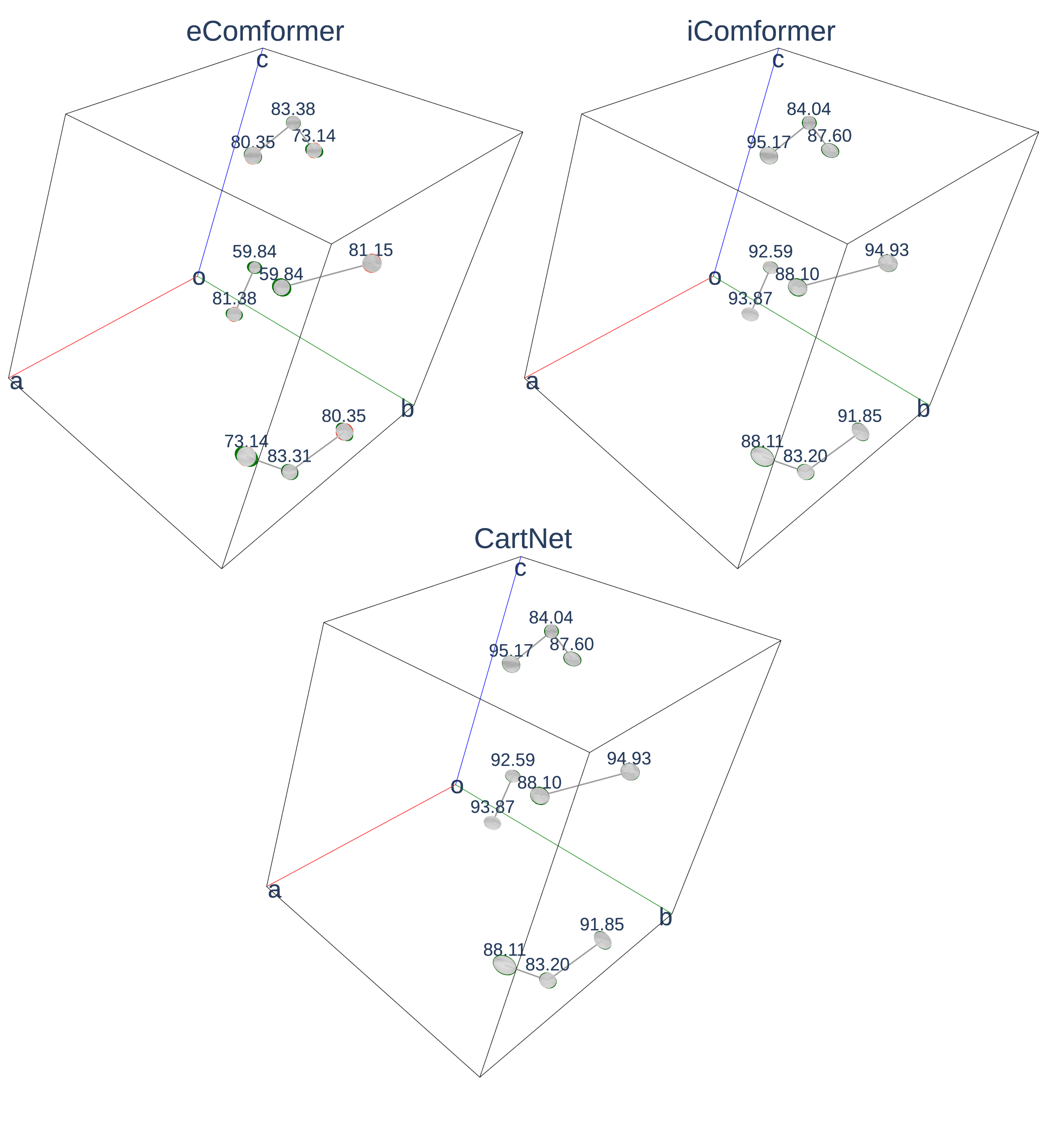}
    \caption{Thermal ellipsoids representations from experimental ADPs for the \textit{glycine} crystal structure (CSD refcode: DOLBIR21) predicted with CartNet, eComformer and iComformer. The green regions represent the experimental values, the red ones represent the prediction values, and the grey represents the intersection between them. The numbers in each atom represent the IoU between the experimental and the calculated ADP.
    Highlighted can be seen a sample ellipsoid predicted using the DFT and the same ellipsoid using CartNet. 
    The parallelepiped represents the unit cell, and the red, green, and blue lines correspond to the a, b, and c unit cell axes. }
    \label{fig:enter-label}
\end{figure}

\subsection{OXALAC10}

\begin{figure}[H]
    \centering
    \includegraphics[width=\linewidth]{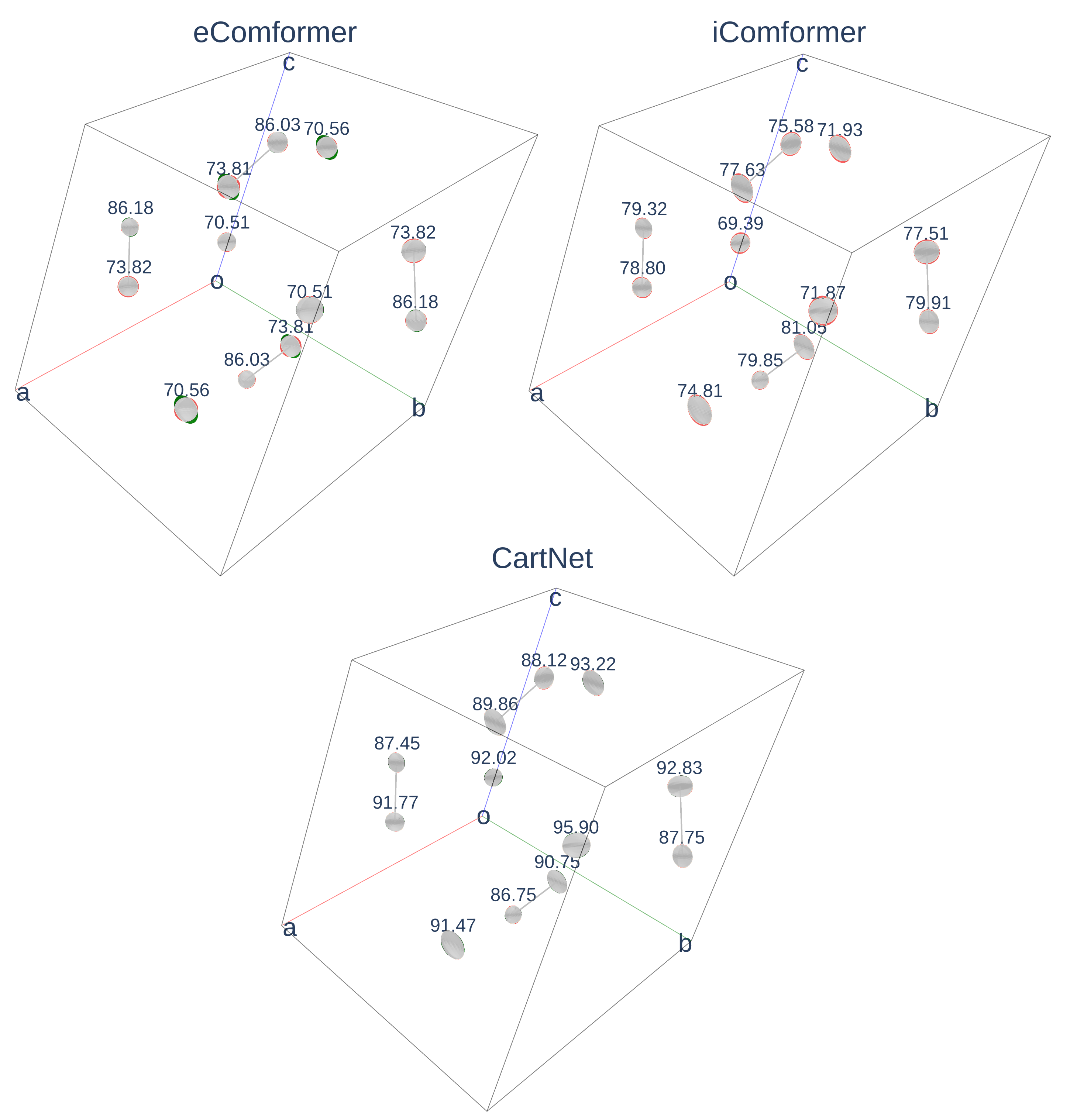}
    \caption{Thermal ellipsoids representations from experimental ADPs for the \textit{Ethane-1,2-dioic acid} crystal structure (CSD refcode: OXALAC10) predicted with CartNet, eComformer and iComformer. The green regions represent the experimental values, the red ones represent the prediction values, and the grey represents the intersection between them. The numbers in each atom represent the IoU between the experimental and the calculated ADP.
    Highlighted can be seen a sample ellipsoid predicted using the DFT and the same ellipsoid using CartNet. 
    The parallelepiped represents the unit cell, and the red, green, and blue lines correspond to the a, b, and c unit cell axes. }
    \label{fig:enter-label}
\end{figure}

\subsection{ITIZOA01}

\begin{figure}[H]
    \centering
    \includegraphics[width=\linewidth]{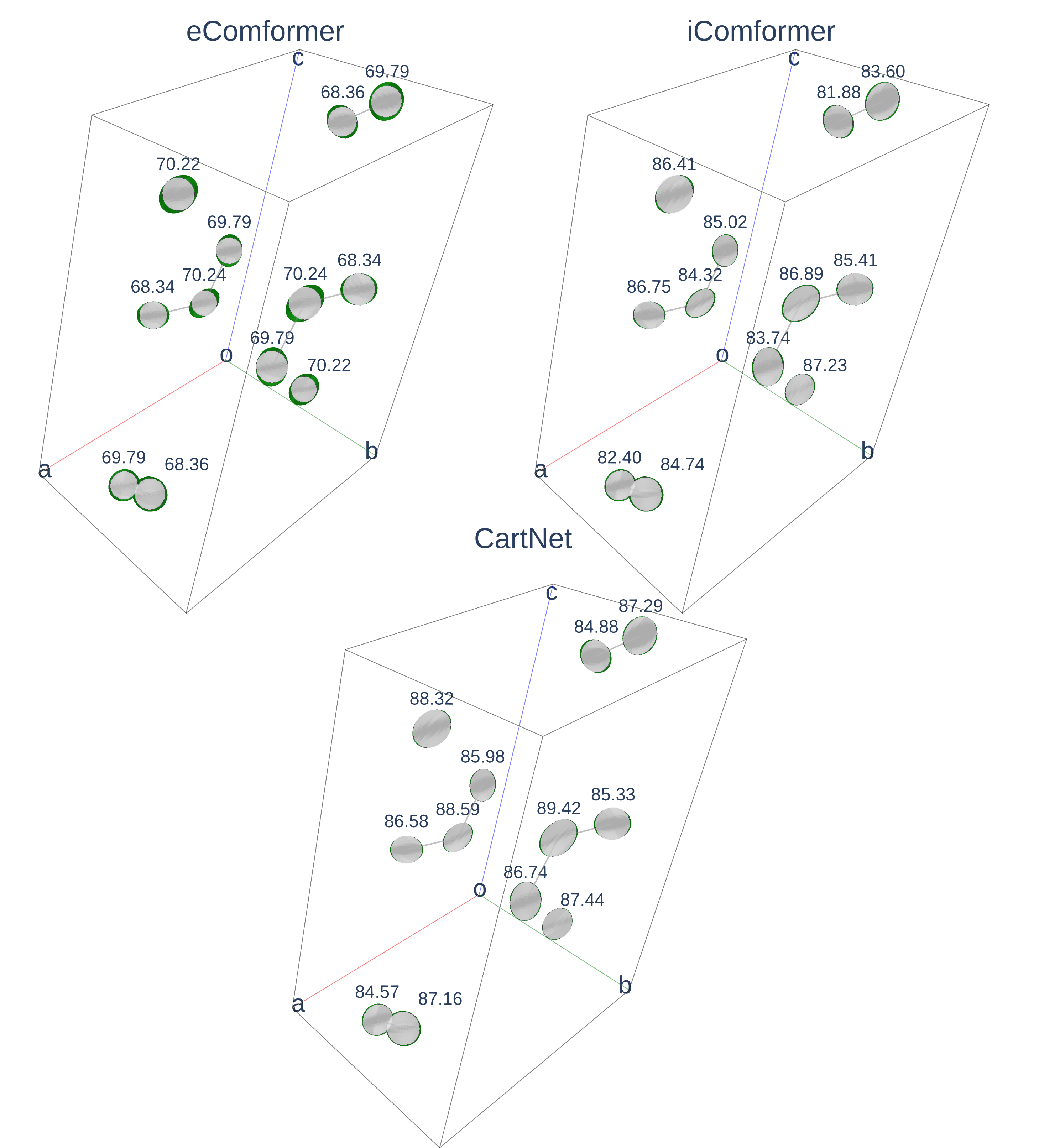}
    \caption{Thermal ellipsoids representations from experimental ADPs for the \textit{piperazine} crystal structure (CSD refcode: ITIZOA01) predicted with CartNet, eComformer and iComformer. The green regions represent the experimental values, the red ones represent the prediction values, and the grey represents the intersection between them. The numbers in each atom represent the IoU between the experimental and the calculated ADP.
    Highlighted can be seen a sample ellipsoid predicted using the DFT and the same ellipsoid using CartNet. 
    The parallelepiped represents the unit cell, and the red, green, and blue lines correspond to the a, b, and c unit cell axes. }
    \label{fig:enter-label}
\end{figure}


\end{document}